%% file: thesis.tex
\PassOptionsToPackage{square,numbers}{natbib}
\documentclass[twoside,11pt]{Latex/Classes/PhDthesisPSnPDF}

\usepackage[T1]{fontenc}
\usepackage{array}
\usepackage{pdfpages}
\usepackage{xspace}
\usepackage{xcolor}
\usepackage{lipsum}
\usepackage{amsmath}
\usepackage{booktabs}
\usepackage{enumitem}
\usepackage{amssymb}
\usepackage{mathrsfs}
\usepackage{multirow}
\usepackage{colortbl} 
\usepackage{makecell}
\setlist[itemize]{topsep=0pt, itemsep=0pt, parsep=0pt, partopsep=0pt}
\setlist[enumerate]{topsep=0pt, itemsep=0pt, parsep=0pt, partopsep=0pt}

\usepackage{graphicx}

\begin{document}

\thispagestyle{empty}

\begin{center}

Vrije Universiteit Amsterdam \hspace*{1.5cm} Universiteit van Amsterdam

\hspace*{-2.6cm}\includegraphics[height=20mm]{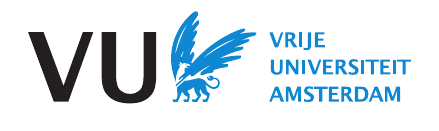}
\hspace*{3.2cm}\includegraphics[height=15mm]{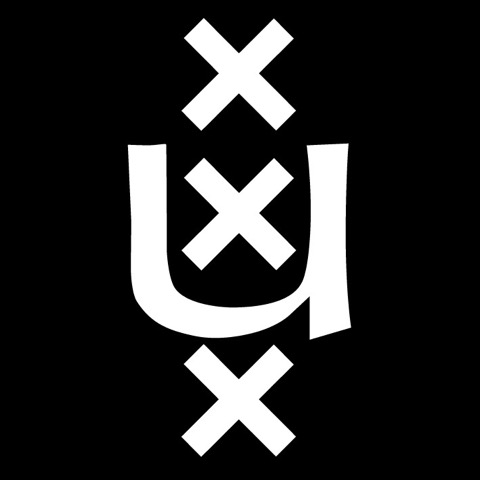}

\vspace{2cm}

{\Large Master Thesis}

\vspace*{1.5cm}

\rule{.9\linewidth}{.6pt}\\[0.4cm]
{\huge \bfseries Towards Adaptive Deep Learning: Model Elasticity via Prune-and-Grow CNN Architectures\par}\vspace{0.4cm}
\rule{.9\linewidth}{.6pt}\\[1.5cm]

\vspace*{2mm}

{\Large
\begin{tabular}{l}
{\bf Author:} ~~Pooja Mangal ~~~~ (2806747)
\end{tabular}
}

\vspace*{2cm}

\begin{tabular}{ll}
{\it University supervisor:}   & ~~Dolly Sapra \\
{\it Daily supervisor:} & ~~Sudaksh Kalra \\
{\it 2nd reader:}       & ~~Emma Beauxis Aussalet 
\end{tabular}

\vspace*{2.5cm}

\textit{A thesis submitted in fulfillment of the requirements for\\ the joint UvA-VU Master of Science degree in Computer Science}

\vspace*{1.8cm}

\today\\[4cm] 

\end{center}

\newpage

       
\hbadness=10000
\hfuzz=50pt



\renewcommand\baselinestretch{1.0}
\baselineskip=16pt plus1pt



\newpage
%
%
%
%




\include{0_frontmatter/abstract}



\frontmatter
\include{0_frontmatter/dedication}
\include{0_frontmatter/acknowledgement}


\setcounter{secnumdepth}{3} 
\setcounter{tocdepth}{3}    
\tableofcontents            


\listoffigures	

\listoftables  


%
\include{0_frontmatter/glossary}

\begin{footnotesize} 

\printnomenclature[1.5cm] 
\label{nom} 

\end{footnotesize}


\mainmatter

\renewcommand{\chaptername}{} 



\input{sections/introduction}
\input{sections/background}

\input{sections/methodology}

\input{sections/experiments}
\input{sections/evaluation}

\input{sections/discussion}

\input{sections/conclusion}






\bibliographystyle{Latex/Classes/PhDbiblio-url2} 
\renewcommand{\bibname}{References} 

\bibliography{references} 










\end{document}

%% file: 0_frontmatter/abstract.tex
\begin{abstracts}
Deploying deep convolutional neural networks (CNNs) on resource-constrained devices presents significant challenges due to their high computational demands and rigid, static architectures. To overcome these limitations, this thesis explores methods for enabling CNNs to dynamically adjust their computational complexity based on available hardware resources. We introduce adaptive CNN architectures capable of scaling their capacity at runtime, thus efficiently balancing performance and resource utilization.

To achieve this adaptability, we propose a structured pruning and dynamic reconstruction approach that creates nested subnetworks within a single CNN model. This approach allows the network to dynamically switch between compact and full-sized configurations without retraining, making it suitable for deployment across varying hardware platforms. Experiments conducted across multiple CNN architectures including VGG-16, AlexNet, ResNet-20, and ResNet-56 on CIFAR-10 and Imagenette datasets demonstrate that adaptive models effectively maintain or even enhance performance under varying computational constraints. Our results highlight that embedding adaptability directly into CNN architectures significantly improves their robustness and flexibility, paving the way for efficient real-world deployment in diverse computational environments.
\end{abstracts}

%% file: 0_frontmatter/dedication.tex





%% file: 0_frontmatter/acknowledgement.tex




%% file: 0_frontmatter/glossary.tex




\nomenclature[A1]{AlexNet}{An early convolutional neural network (CNN) architecture known for its success in image classification tasks.}
\nomenclature[A2]{VGG-16}{A deep convolutional network known for its uniform structure and commonly used as a benchmark in image classification tasks.}
\nomenclature[A3]{ResNet}{Residual Neural Network; a deep learning architecture using residual connections.}

\nomenclature[B1]{Pruning}{The process of removing unnecessary weights, channels, or layers from a neural network to reduce its size and computational cost.}
\nomenclature[B2]{Core Model}{The smallest version of the original model obtained after pruning, containing only the important (unpruned) parameters.}
\nomenclature[B3]{Rebuilding}{The process of restoring or reconstructing pruned components of a neural network—such as filters, channels, or layers—back into the model architecture to recover capacity, accuracy, or adapt to improved resource availability at runtime.}
\nomenclature[B4]{Dependency Graph}{A graph structure used to model inter-layer dependencies in neural networks, ensuring valid architecture during structured pruning.}

\nomenclature[C1]{Magnitude Importance}{A pruning importance score that ranks parameters based on the magnitude of their weights, assuming that smaller-magnitude weights contribute less to the model’s output.}
\nomenclature[C2]{Hessian Importance}{A pruning importance score that uses second-order derivatives (Hessian) to measure how sensitive a parameter is to the loss function.}
\nomenclature[C3]{Taylor Importance}{A pruning method that uses first-order Taylor series expansion to estimate the effect of removing a channel on the loss function.}

\nomenclature[D1]{Cross-Entropy Loss}{A common loss function used for classification problems that measures the difference between predicted and true class distributions.}

%% file: sections/introduction.tex
\chapter{Introduction}\label{ch:introduction}

Deep neural networks (DNNs), particularly convolutional neural networks (CNNs), have become foundational in modern computer vision tasks, achieving state-of-the-art performance in areas ranging from object detection to scene understanding. However, these performance gains often come at the cost of increased computational demands, making it challenging to deploy such models on resource-constrained or dynamically varying platforms like mobile devices, drones, embedded sensors, and edge computing nodes. These environments often have limited and fluctuating resources such as CPU cycles, memory, and battery power, rendering the deployment of static, monolithic CNNs is inefficient. For example, ResNet with 152 layers requires 11.3G MAC operations to achieve its high inference accuracy. This tremendous computational cost poses challenges when DNNs are applied in resource-constrained and resource-varying platforms, e.g., mobile phones and autonomous vehicles. \cite{SteppingNet}

To address these challenges, there has been growing interest in adaptive neural networks that can scale their computational footprint based on the availability of system resources. Such networks exhibit model elasticity, the ability to adapt their structure, or execution dynamically to balance accuracy and efficiency. Previous work in this area \cite{NestDNN, legoDNN, slimmableneuralnetworks} introduces strategies to make deep models more flexible and deployable in real world settings. These methods often rely on pre-computed multi-capacity models, block-wise modularity, or conditional execution to provide runtime adaptability.

\textbf{Challenges and our Solutions:} Building neural networks that can adapt to changing resource availability in real time comes with two main challenges. First, most existing pruning methods create a fixed version of the model that cannot be adjusted once deployed. After pruning, parts of the model are permanently removed, which means the network cannot regain its original capacity when more resources become available. One solution is to deploy multiple versions of the model, each pruned to a different size, and switch between them based on system load. However, this approach is not scalable, especially when running several models at once, as it consumes too much memory and introduces high switching overheads for paging in and paging out complete models. The second challenge lies in reconstructing the pruned model accurately. When layers or channels are removed, rebuilding them requires careful tracking of tensor shapes, parameter indices, and layer dependencies to avoid mismatches and maintain performance. Without proper alignment or retraining, reintroducing filters or connections can lead to inconsistent behavior or poor accuracy leading to drops in performance.

To address the first challenge, we implement a flexible pruning and rebuilding approach that creates a single model with single or multiple “nested” versions inside it. Smaller models are embedded within larger ones by sharing weights, so they can be activated or expanded depending on the available resources. When the system is under load, a smaller version of the network could run to save computation. When resources are available again, previously pruned parts can be restored to regain full performance. This allows the model to scale up or down without {adding additional computation and switching overheads.
To address the second challenge, we design a flexible regrowth mechanism that restores pruned weights and layers while preserving the original structure and compatibility of the model. By carefully handling parameter alignment, indexing, and layer dependencies, the network can be rebuilt without introducing inconsistencies or large gaps in performance.

\subsection{Research Questions}
To overcome the limitations described above, this thesis is guided by the following research questions:
\begin{enumerate}
    \item \textbf{RQ1:} How effective are dependency-graph-based structured pruning techniques for achieving efficient and reliable model compression in convolutional neural networks?

    \item \textbf{RQ2:} How can dynamic layer freezing and reconstruction techniques be leveraged to achieve runtime adaptivity of CNNs under constrained and dynamically changing resource environments?
    
    \item \textbf{RQ3:} How can we build nested CNN models that dynamically choose the right size network to balance accuracy and speed on edge devices?

\end{enumerate}

\subsection{Proposed Method}

The goal of this thesis is to design and implement a convolutional neural network (CNN) that exhibits model elasticity: the ability to scale its computational complexity up or down based on changing resource availability. Our method achieves this by applying structured pruning using Dependency Graphs \cite{depGraph, torchpruning} to remove redundant components of the network and then rebuilding them, enabling dynamic transitions between compact and full-capacity versions of the model.

We start with a trained baseline CNN and apply structured pruning at the filter level, targeting convolutional layers that have a low contribution to the final output. To ensure that pruning preserves the functional integrity of the model, we use dependency-aware pruning through the DepGraph utility provided by Torch-Pruning \cite{depGraph}. DepGraph builds a graph that shows how the layers in the model are connected and depend on each other. This helps us prune the model in a way that automatically updates all related layers, so the structure stays valid and the model still functions properly after pruning. Unlike conventional approaches where pruning is final, our method preserves the necessary structural and indexing information to allow for regrowth.

The pruning step produces a smaller, efficient sub-network called the \textit{core model}. During the rebuilding phase, we restore previously pruned filters and their associated weights. This is done while maintaining alignment with the original architecture by ensuring correct indexing and tensor shape matching. This regrowth mechanism supports smooth transitions between model sizes, allowing the network to regain capacity and improve performance without retraining from scratch.

The final outcome is a nested architecture in which multiple capacity levels are embedded within a single model. This design supports elasticity, enabling the network to adapt to different resource constraints by scaling up or down as needed. 

The effectiveness of this approach is validated on CIFAR-10 dataset and various architectures like AlexNet, ResNet20, ResNet56 and VGG-16, demonstrating that it reduces model size and computational requirements while maintaining accuracy.

This thesis is organized as follows: Chapter~\ref{ch:background}  reviews related literature on pruning methods, adaptivity in neural networks, and dependency graph pruning. Chapter~\ref{ch:methodology} explains the methodology, including one-shot and iterative pruning techniques, and rebuilding strategy. Chapter~\ref{ch:experiments} presents the experimental results, highlighting key findings and insights. Chapter~\ref{ch:evaluation} evaluates the model's performance using various metrics and comparisons. Chapter~\ref{ch:discussion} discusses the findings in relation to the research questions and overall objectives. Finally, Chapter~\ref{ch:conclusion} summarizes the contributions, limitations, and suggestions for future research.

%% file: sections/background.tex
\chapter{Background and Related Work}\label{ch:background}

This chapter provides the necessary background for understanding the design and motivation behind adaptive neural networks. It covers the fundamentals of CNNs, the concept of model elasticity, and the role of dependency-aware pruning. It also reviews existing models and methodologies along with background information with a detailed literature review.

\section{Background}

\subsection{Convolutional Neural Networks}

Convolutional Neural Networks (CNNs) are widely used in computer vision tasks such as image classification, object detection, and segmentation. Their success lies in their ability to automatically learn spatial hierarchies of features through convolutional filters. As the demand for higher accuracy grows, CNN architectures have become deeper and more complex, with increased parameters and computational cost. While powerful, these models are often too large and compute-heavy for deployment on resource-constrained devices like smartphones, edge processors, or embedded systems.

\subsection{The Need for Model Efficiency}

Deploying deep models on devices with limited resources requires reducing their size and computational complexity without significantly sacrificing performance. Several model compression techniques have been developed to address this issue, including weight quantization, knowledge distillation, and pruning \cite{deepcompressioncompressingdeep, pruningfiltersefficientconvnets, importanceEstimationForNNPruning, learningefficientconvolutionalnetworks, learningstructuredsparsitydeep, depGraph}. Among these, structured pruning has emerged as a practical solution. It removes entire filters, channels, or layers, leading to smaller models. However, most pruning techniques result in a permanently compressed model, lacking flexibility once deployed.

\subsection{Adaptivity and Model Elasticity}

To make CNNs more suitable for real-world deployment, researchers have proposed the idea of model elasticity \cite{slimmableneuralnetworks, SteppingNet, NestDNN, legoDNN, onceForAll}. An elastic model can dynamically adjust its size and computational cost depending on the runtime resource availability. This means that when more resources are available (e.g., high battery, low CPU load), the model can operate at full capacity for better accuracy. In contrast, when resources are constrained, a smaller version of the model can be used to maintain responsiveness. Adaptive models such as NestDNN, Slimmable Networks, and LegoDNN \cite{NestDNN, legoDNN, slimmableneuralnetworks} have shown how this idea can improve the performance and efficiency of deep learning applications across different conditions and platforms.

\subsection{Dependency-Aware Pruning with DepGraph}

Structured pruning is carried out using DepGraph, a dependency modeling approach, through its implementation in the Torch-Pruning library. DepGraph constructs a computational graph of the model and identifies the dependencies between different layers and parameters. This allows pruning operations, such as filter or channel removal, to be applied in a dependency-aware manner. When a particular layer is pruned, DepGraph automatically updates all related layers to ensure that the overall structure and data flow of the model remain valid. This prevents common issues such as mismatched tensor shapes or broken connections that can arise during manual or uncoordinated pruning. By handling these updates automatically, DepGraph provides and end to end pipeline for structured pruning resulting in compact and lightweight models.

\section{Related Work}

Recent research has focused on designing adaptive neural networks that exhibit model elasticity. This section reviews the major developments in this area, focusing on approaches that align closely with this thesis’s goals of adaptation and runtime model reconstruction.

\subsection{Resource-Aware Adaptive Models}

One of the foundational works in resource-aware model adaptation is NestDNN proposed by Fang et al. \cite{NestDNN}, which introduces a multi-capacity CNN architecture designed for mobile vision tasks. NestDNN constructs a family of nested sub-networks, each representing a different trade-off point between accuracy and computational cost. These sub-networks share weights and are generated via iterative pruning followed by a freeze-and-grow mechanism that incrementally reconstructs the larger network from the smaller seed models. A runtime scheduler dynamically selects which sub-model to execute based on current system conditions, enabling efficient multi-tenant deployment. This concept of progressive capacity scaling and resource-aware scheduling is foundational to the dynamic pruning-and-rebuilding strategy explored in this thesis.

Similarly, Han et al \cite{legoDNN} offers a block-wise modular approach to elasticity. Rather than nesting, LegoDNN pre-trains a compact set of reusable network blocks, which can be assembled at runtime to create models with varying computational footprints. This block-grained scaling strategy provides fine-grained control over the trade-off between efficiency and performance and supports rapid reconfiguration without retraining. LegoDNN emphasizes the composability of neural network modules, highlighting the practical feasibility of runtime model reconstruction in real-world systems.

Once-for-All (OFA) \cite{onceForAll} further generalizes this idea by training a single supernetwork that supports a wide range of architectural configurations including different kernel sizes, channel widths, and depths without retraining. OFA enables zero-cost specialization for diverse hardware targets by sampling subnetworks from the supernetwork, making it highly suited for on-the-fly adaptation in heterogeneous environments. OFA introduces continuous elasticity with minimal latency overhead.

Cheng et al\cite{SteppingNet} proposes an incremental architecture where the base model executes a minimal forward pass and gradually adds layers to improve accuracy based on runtime constraints or prediction confidence. This architecture strikes a balance between early-exit models and monolithic networks, offering a step-wise refinement mechanism for dynamic adaptation. Like the pruning-regrowth concept in this thesis, SteppingNet leverages modular stages that can be conditionally activated, enabling graceful trade-offs between performance and resource consumption.

Slimmable Neural Networks and Universally Slimmable Networks (US-Nets) \cite{slimmableneuralnetworks, universallySlimmableNetworks} enable a single CNN to be executed at various widths (i.e., number of channels per layer) with shared parameters. These models are trained to support multiple predefined or even arbitrary width settings, allowing inference-time adjustment based on available computation budgets. The training approach ensures that each sub-network within the slimmable model retains good predictive performance, offering elasticity without the need for multiple models. This principle of channel-wise scaling is conceptually compatible with pruning-based dynamic adaptation, where the number of active filters or layers can be altered on-the-fly.

MutualNet \cite{mutualnetadaptiveconvnet} further extends this idea by jointly adapting both network width and input resolution, learning a set of sub-networks via mutual learning. This enables simultaneous optimization of architectural complexity and input fidelity. The mutual learning approach ensures better knowledge transfer between sub-models, enhancing the accuracy of each variant and pointing to an effective strategy for managing trade-offs during pruning and regrowth.

\subsection{Pruning Techniques}

Pruning techniques can broadly be categorized into \textit{unstructured} and \textit{structured} pruning. Unstructured pruning removes individual weights from the network based on their magnitude, leading to sparsity in the weight matrices. While this can significantly reduce the number of parameters, it often results in irregular memory access patterns and limited real-world speedups unless specialized sparse hardware or libraries are used.

In contrast, structured pruning removes entire filters, channels, or blocks, leading to a more compact and efficient network architecture that aligns better with conventional hardware. Since this thesis focuses on enabling runtime elasticity i.e., adapting the network to different compute budgets without requiring sparse computation, structured pruning is the more suitable choice. It simplifies model compression and regrowth while ensuring compatibility with off-the-shelf inference engines.

This work builds on several foundational pruning strategies. NestDNN \cite{NestDNN} introduces a freeze-and-grow mechanism that constructs a multi-capacity model by incrementally pruning and rebuilding sub-networks. We adopt a similar strategy by combining both one-shot and iterative structured pruning strategies, along with dynamic reconstruction of pruned channels, to build elastic CNNs.

Other notable works include Pruning Filters for Efficient Convnets \cite{pruningfiltersefficientconvnets}, which proposes L1-norm-based filter pruning, a simple yet effective metric for structured pruning. The Lottery Ticket Hypothesis \cite{lotteryTicket} demonstrates the existence of sub-networks (winning tickets) that can be trained in isolation to match full model performance, providing conceptual insights into pruning and reinitialization. DepGraph\cite{depGraph} improves structured pruning by analyzing dependency graphs between layers, ensuring pruning decisions do not lead to disconnection or invalid topologies. These methods inform both the pruning and regrowth steps necessary for building truly elastic CNNs.

Together, these approaches inform the pruning and regrowth mechanisms used in this thesis, forming the basis for constructing compact yet adaptable neural networks suitable for varying runtime constraints.

%% file: sections/methodology.tex
\chapter{Methodology}\label{ch:methodology}

This section describes the step-by-step procedure followed to prune, fine-tune, and rebuild the neural network to allow for adaptivity. The goal of our approach is to create a flexible method that adjusts the model's resource usage based on available compute resources and the current state of its environment. 

\section{Pruning Strategies}

In this work, we implemented and compared several structured pruning strategies to reduce the size of deep neural networks while preserving accuracy. Our pruning strategy evolved through multiple stages, beginning with basic structured and unstructured pruning techniques and eventually incorporating dependency graph-based pruning. 

Given the complexity of modern architectures such as skip connections in ResNet or concatenation in DenseNet, naïvely removing filters or channels can lead to inconsistencies and runtime errors. To address this, all pruning strategies in this thesis are implemented using the \textit{Dependency Graph (DepGraph)} method introduced by Fang et al. (2023) \cite{depGraph}.

\subsection*{Filter Pruning}

\begin{figure}[ht]
    \centering
    \includegraphics[width=0.9\linewidth]{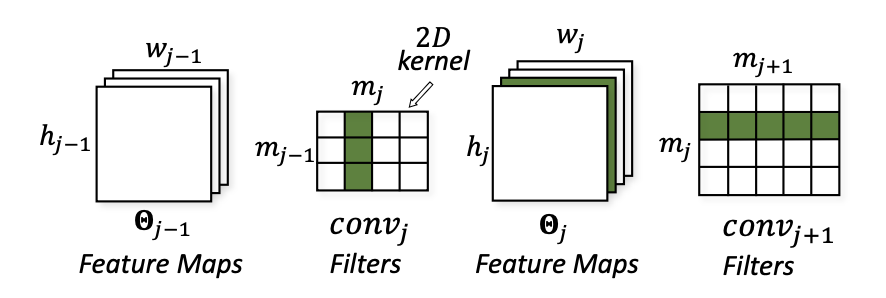}
    \caption{Overview of the iterative pruning and rebuilding process. The model is progressively compressed via structured pruning steps and later reconstructed in reverse using preserved weight metadata. \cite{pruningfiltersefficientconvnets}}
    \label{filter_pruning}
\end{figure}

Figure \ref{filter_pruning} illustrates the details of filter pruning. Let $\Theta_{j-1} \in \mathbb{R}^{w_{j-1} \times h_{j-1} \times m_{j-1}}$ denote the input feature maps of the $j$th convolutional layer $conv_j$ of a CNN, where $w_{j-1}$ and $h_{j-1}$ are the width and height of each of the input feature maps; and $m_{j-1}$ is the total number of the input feature maps. The convolutional layer $conv_j$ consists of $m_j$ 3D filters with size $k \times k \times m_{j-1}$ ($k \times k$ is the 2D kernel). It applies these filters onto the input feature maps $\Theta_{j-1}$ to generate the output feature maps $\Theta_j \in \mathbb{R}^{w_j \times h_j \times m_j}$, where one 3D filter generates one output feature map. This process involves a total of $m_j k^2 m_{j-1} w_j h_j$ floating point operations (i.e., FLOPs).

Since one 3D filter generates one output feature map, pruning one 3D filter in $conv_j$ (marked in green in $conv_j$) results in removing one output feature map in $\Theta_j$ (marked in green in $\Theta_j$), which leads to $k^2 m_{j-1}$ parameter and $k^2 m_{j-1} w_j h_j$ FLOPs reduction. Subsequently, $m_{j+1}$ 2D kernels applied onto that removed output feature map in the convolutional layer $conv_{j+1}$ (marked in green in $conv_{j+1}$) are also removed. This leads to an additional $k^2 m_{j+1}$ parameter and $k^2 m_{j+1} w_{j+1} h_{j+1}$ FLOPs reduction. Therefore, by pruning filters, both model size (i.e., model parameters) and computational cost (i.e., FLOPs) are reduced~\cite{pruningfiltersefficientconvnets}.

\subsection*{Handling Dependencies with Dependency Graph Pruning}As neural networks become more interconnected, pruning decisions made in one layer can affect other layers. For instance, in a ResNet block, reducing the number of output channels in one convolution requires matching those changes in the residual branch to maintain dimension compatibility.

\begin{figure}[ht]
    \centering
    \includegraphics[width=\linewidth]{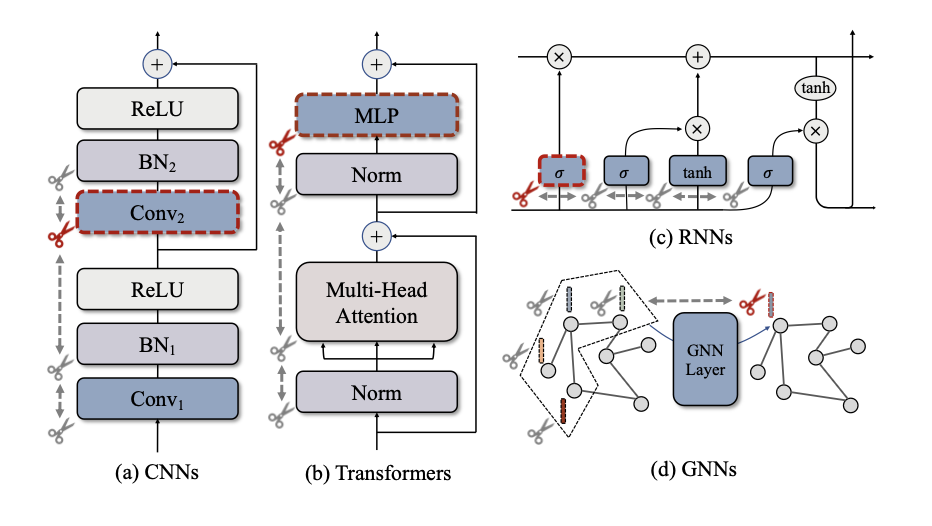}
    \caption{Dependency-aware pruning ensures consistency across layers with structural dependencies. For example, pruning Conv2 also requires pruning Conv1, BN1, and BN2 within the block. The Dependency Graph tracks such relationships automatically. Source: \cite{depGraph}}
    \label{fig:depgraph}
\end{figure}

DepGraph can find groups of layers that must be pruned together. This enables safe, automatic pruning by ensuring that structurally coupled layers are pruned together. Figure~\ref{fig:depgraph} illustrates how pruning one layer can require simultaneous changes in others to preserve network validity.

In all experiments, DepGraph is used as the backbone for pruning, allowing us to experiment with different filter selection criteria and pruning configurations, without compromising structural integrity.

\subsection{Manual Layer-Wise Pruning using Dependency Graphs}
This approach uses Torch-Pruning's \textit{Dependency Graph (DepGraph)} to ensure structural consistency during pruning. In this method:
\begin{itemize}
    \item Specific convolutional layers are manually selected for pruning.
    \item For each selected layer, the \textbf{L1 norm} of each filter (i.e., output channel) is computed to estimate its importance.
    \item A fixed percentage of the least important filters is pruned from each selected layer independently, making the pruning selection local.   
    \item The Dependency Graph tracks all parameter dependencies and ensures that pruning one layer automatically updates all structurally coupled layers (e.g., connected BatchNorm or downstream layers).
\end{itemize}

This method offers fine-grained control and transparency, making it suitable for targeted experiments or analysis. However, it does not adapt pruning ratios across layers and may prune equally from both redundant and critical layers. It is also difficult to extract all the indices that are being pruned using this method. This method also limits scalability and automation, as it requires manual selection of layers and pruning indices.

\vspace{0.5em}

\subsection{High-Level Pruning Using Magnitude Importance}

This strategy uses Torch-Pruning's built-in \texttt{MagnitudePruner} and \texttt{MagnitudeImportance} modules to automate pruning decisions across the entire model.

\begin{itemize}
    \item \textbf{Layer selection is automatic:} The pruner evaluates all convolutional layers, excluding only user-defined exceptions, such as the final classification layer.
    \item \textbf{Importance is computed globally:} It calculates the \textbf{L2 norm} (Euclidean norm) across all layers to determine their relative importance in the network.
    \item \textbf{Pruning is applied globally:} The pruning ratio is applied throughout the model, allowing the pruner to prune more from redundant layers and less from sensitive ones.
    \item \textbf{Structural consistency is preserved:} Internal use of \texttt{DependencyGraph} ensures the network structure remains valid after pruning.
\end{itemize}

\vspace{0.5em}

\noindent\textbf{Key Difference:}  
\textit{Manual pruning applies a fixed ratio independently to selected layers, while high-level pruning evaluates importance across all layers and adapts pruning accordingly, often resulting in better performance.}

\subsection{High-Level Pruning Using Taylor Importance}
In this method, importance is determined using the \textbf{Taylor expansion}:
\begin{itemize}
    \item It estimates the effect of pruning on the loss function using gradients.
    \item A representative batch of training data is used to compute the gradients.
    \item The least important filters are pruned on the basis of their gradient impact on the loss.
\end{itemize}

Taylor-based pruning is more accurate than magnitude-based methods because it takes the model's loss landscape into account.

\subsection{High-Level Pruning Using Hessian Information}
This technique uses second-order derivatives (Hessian) to evaluate channel importance:
\begin{itemize}
    \item It approximates the \textbf{Hessian matrix} to assess sensitivity of each parameter.
    \item Several mini-batches are used to get stable Hessian estimates.
    \item The Hessian values are scaled to improve discrimination between important and unimportant filters.
\end{itemize}

Although more computationally intensive, this approach yields highly informed pruning decisions.

\subsection{Soft Pruning (Zeroing Out Weights)}
In soft pruning, weights are not removed but set to zero:
\begin{itemize}
    \item Magnitude-based criteria are used to identify unimportant channels.
    \item Weights of those filters are multiplied by zero instead of being deleted.
    \item This maintains the original architecture and avoids the need for structural changes.
\end{itemize}

Soft pruning is useful when model compatibility and shape preservation are a priority.

These strategies were evaluated using VGG-16, AlexNet, ResNet20 and ResNet56 on the CIFAR-10 dataset to understand their effect on performance and adaptability.

\begin{table}[ht]
\centering
\caption{Comparison of Pruning Strategies Used in This Work}
\label{tab:pruning-strategies}
\begin{tabular}{|p{3.5cm}|p{4.5cm}|p{3.5cm}|p{3cm}|}
\hline
\textbf{Pruning Strategy} & \textbf{Importance Metric} & \textbf{Advantages} & \textbf{Challenges} \\
\hline
\textbf{Manual layer-wise pruning} & L1 norm & Preserves structural integrity, handles inter-layer dependencies & Manual layer selection required \\
\hline
\textbf{High-Level Magnitude Pruning} & L2 norm & Fully automated, efficient across layers & Ignores gradient information \\
\hline
\textbf{High-Level Taylor Pruning} & First-order Taylor expansion (gradient × activation) & Considers impact on loss, better accuracy retention & Requires gradient computation \\
\hline
\textbf{High-Level Hessian Pruning} & Second-order derivatives (Hessian approximation) & Most informed, loss-sensitive pruning decisions & High computation cost, memory intensive \\
\hline
\textbf{Soft Pruning} & L2 norm & Maintains model shape, avoids reconfiguration & Weights are zeroed but not removed, limited speedup \\
\hline
\end{tabular}
\end{table}

\section{Core Model Generation (Pruned Model)}
After pruning, the next step was to construct the core model, a smaller version of the original architecture that retains only the unpruned channels. For most methods (e.g., manual layer-wise pruning, magnitude-based, Taylor, or Hessian-based pruning), this step is handled automatically by the library. The pruning framework physically removes pruned filters and updates the model structure accordingly. The resulting pruned model can be directly used for further training or evaluation.

However, in the case of soft pruning, the pruned weights are not removed but set to zero. This means the model still includes inactive channels that increase memory usage and computational cost. To address this, we used a custom reconstruction process: The pruned model is first analysed to look at the number and position of active (unpruned) channels in each layer to check which channels removed (i.e., have zero-valued weights across all filters). A new model is initialized using only the retained channels, effectively reducing its size and complexity. The corresponding weights from the original pruned model are then copied into this new model, excluding any zeroed-out or pruned weights. This reconstruction strategy is applied uniformly across all models in this thesis.

\section{Rebuilding the Full Model for Adaptivity}

Once we get to our core model, the next step is rebuilding a full-sized model that structurally resembles the original network but contains both pruned and unpruned components. This reconstruction step plays an important role in enabling adaptive deployment of the model across varying computational environments.
\subsubsection*{Why Rebuild the Model?}
After pruning, we rebuild the model to enable adaptivity. Instead of creating and switching between multiple separate models, rebuilding helps us construct a single model that contains multiple descendant models having smaller, pruned versions nested within it. This rebuilt model maintains a structure where different layers or blocks can be swapped or resized efficiently, much like Lego bricks in LegoDNN \cite{legoDNN}. For example, on a powerful device, the full model can run for maximum accuracy, while on a mobile phone with low battery, a smaller sub-model can be activated instantly to save energy. Rebuilding enables flexibility by ensuring all the internal connections and parameter shapes are correctly aligned after pruning so different parts can be reassembled as needed without retraining or reinitializing the model every time. It turns a static model into a dynamic, responsive one.

To addresses this, we reintroduce the pruned channels or weights into the architecture in their original positions, using the original model parameters. This allows the model
to regain its full capacity while retaining the structural benefits of pruning. Moreover,
we fine-tune the rebuilt model, allowing the network to adapt to the new / reinserted
components and potentially achieve higher performance than the pruned model.

\subsubsection*{How Rebuilding Works}

The rebuilding process consists of four main steps:

\paragraph{1. Identifying pruned and unpruned components.} From the pruning phase, we collect detailed information about each layer, including: \begin{itemize} \item Which channels were retained (unpruned) and their corresponding weights. \item Which channels were removed (pruned), along with their original positions in the network. \item Using the information from the pruned core model, we compute the number of channels that were pruned and add them back to create a channel configuration matching the original model’s size. This defines the structure of the rebuilt model. \end{itemize}

\paragraph{2. Initializing a full-sized model.} Using the original model architecture as a reference, we initialize a new model with the same number of layers and channels. At this point, the model has no weights assigned but has the same structure as the original unpruned network.

\paragraph{3. Reconstructing weights.} We populate the weights of the rebuilt model using our saved information from the pruning step: \begin{itemize} \item The weights of the unpruned channels are inserted into their correct positions. \item Pruned weights are placed back into the model as well, so that the network structure is complete. \end{itemize} This results in a model that contains all channels (active and inactive) and whose weights are fully restored based on the pruning step.

\paragraph{4. Freezing Channels:} 
At each step of the rebuilding phase, we carefully freeze the unpruned (core) channels i.e., the weights retained from the previous level of pruning. This means that the important features the model has already learned stay untouched while we fine-tune only the newly added channels.

This step is essential for enabling adaptability across multiple resource configurations. Without freezing, each time we rebuild and fine-tune, the earlier learned weights would be changed, creating inconsistency between different model versions. By preserving the core weights across rebuild levels, we ensure that the final model contains a reusable subnetwork corresponding to the original pruned core. This strategy enables runtime switching between models of different capacities, making the network elastic and efficient across diverse deployment settings.

\begin{figure}[ht]
    \centering
    \includegraphics[width=\linewidth]{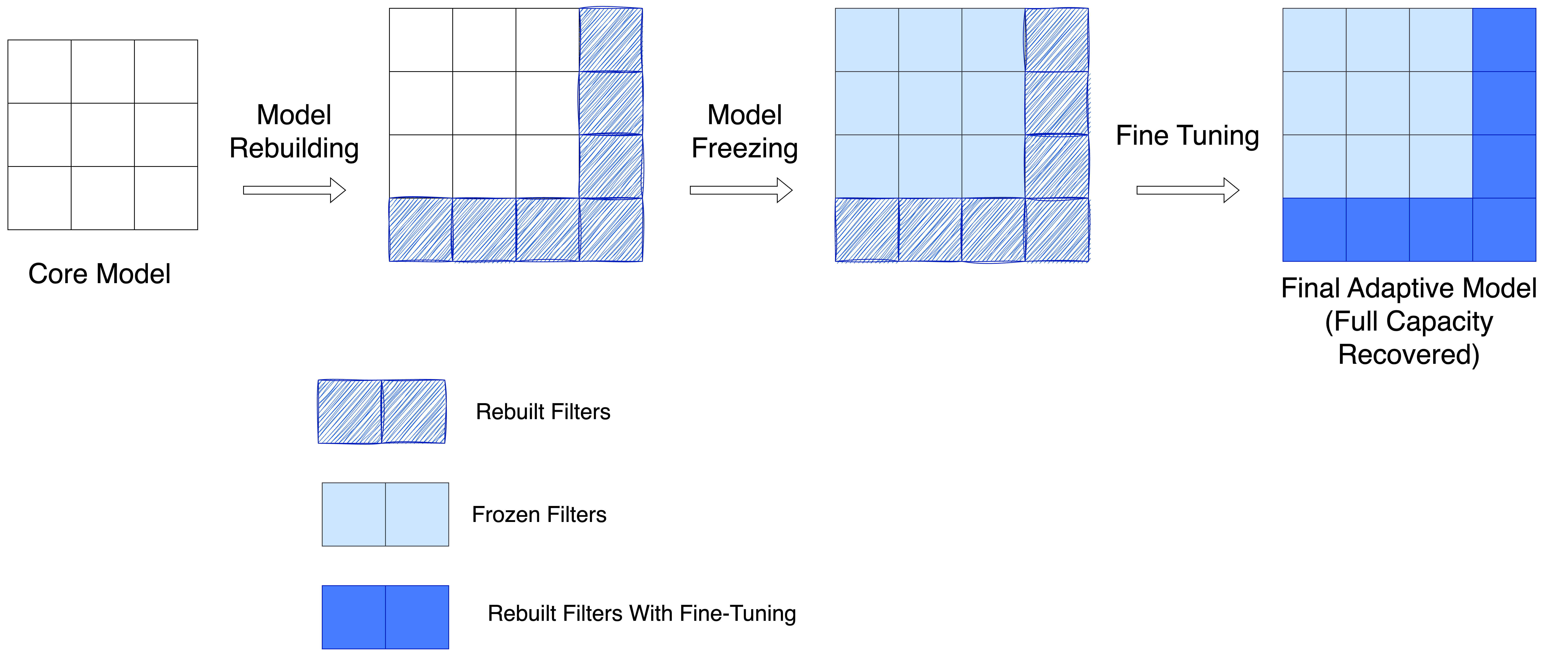}
    \caption{Illustration of model freezing and rebuilding}
    \label{fig:model_freezing}
\end{figure}

Figure \ref{fig:model_freezing} illustrates the process of recovering a larger model from a compact core model using freezing and rebuilding strategy. Starting from the pruned core model, previously removed filters are reinserted and initialized with their original weights. The existing filters are frozen, and their parameters remain unchanged during training. Only the reintroduced filters are updated through fine-tuning. This results in a final model that embeds the original core model while expanding its capacity, making it adaptable to different resource budgets.

\begin{figure}[ht]
    \centering
    \includegraphics[width=\linewidth]{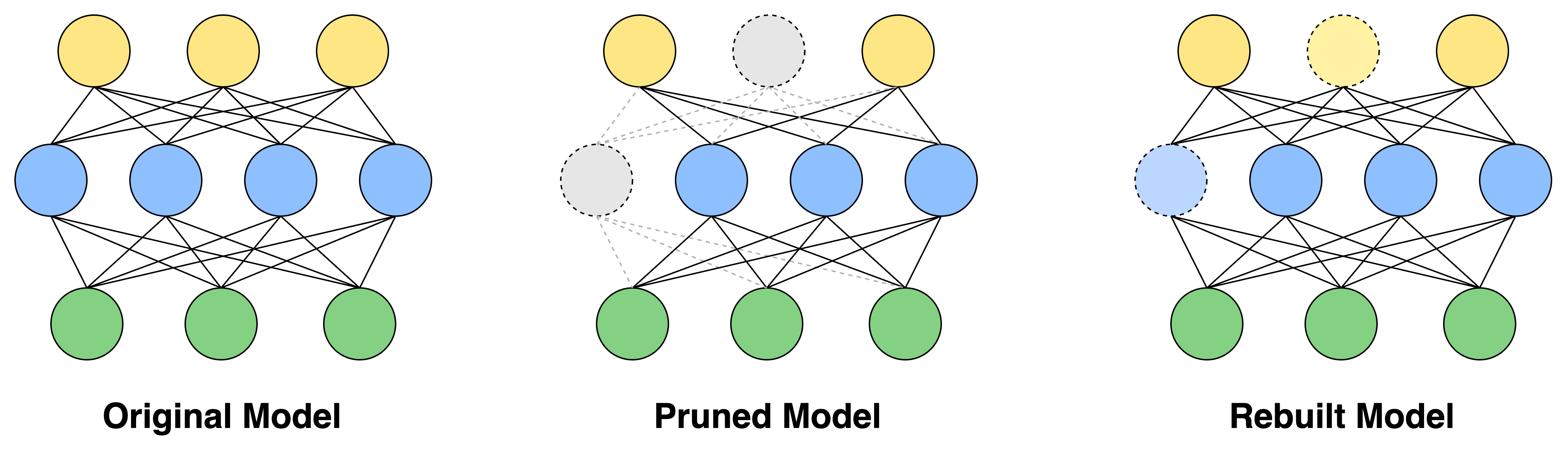}
    \caption{Overview of the pruning and rebuilding process for adaptivity. Structured pruning removes entire filters (shown as dotted circles), resulting in a smaller core model. Later, previously pruned filters are reinserted into the architecture during the rebuilding phase, allowing the model to adapt to increased resource availability.}
    \label{fig:prune-rebuild-diagram}
\end{figure}
Figure~\ref{fig:prune-rebuild-diagram} illustrates the overall process of pruning and rebuilding. The original model is first pruned to remove entire filters, resulting in a smaller, more efficient core model. These pruned filters are not discarded, instead, their weights and positions are stored for later reconstruction. When more resources become available, the model can be rebuilt by reinserting the previously removed filters, forming a larger, adaptive version of the network.

\section{Fine-Tuning Process}
After pruning and rebuilding the model, fine-tuning is a crucial step to recover accuracy. Since pruning removes parts of the network, and rebuilding alters its structure, the model needs to synchronize the removed (Pruning) or new (Rebuilding) connections for the classification tasks. Fine-tuning allows the network to adjust its weights and adapt to its new structure, whether it is the smaller pruned model or the rebuilt, full-size version.

We perform two rounds of fine-tuning in this process: one before the rebuilding stage and another after rebuilding.

\subsubsection*{Fine-Tuning Before Rebuilding}

Once the model has been pruned, and we achieve our final core model, its size is reduced, and some of its parameters are removed. This leads to a drop in performance, as some important channels might be removed. To address this, we fine-tune the core model on the original training dataset for different number of epochs. This helps the remaining channels learn to compensate for the missing ones, and improves overall performance without increasing model size.

Throughout training, we track metrics such as training loss, validation loss, accuracy, and learning rate progression. These are logged using a monitoring platform for better analysis and reproducibility.

\subsubsection*{Fine-Tuning After Rebuilding (Adaptive Model)}
After rebuilding the full-sized model, a second round of fine-tuning is performed to optimize its performance. The rebuilt model contains both the retained channels from the pruned (core) model and the reinserted channels that were previously removed. While the retained channels already carry learned knowledge, the reinserted channels also come from the original pre-trained model and thus contain useful initial knowledge. This strong initialization helps reduce the number of fine-tuning epochs required since the training primarily focuses on reconnecting and integrating these restored channels effectively with the existing ones.

To handle this properly, we apply a selective fine-tuning approach. The main idea is to train only the newly added parameters while freezing the important parts learned during pruning. This is achieved by explicitly zeroing out the gradients of the retained (unpruned) channels during backpropagation. As a result, the optimizer updates only the reinserted channels, allowing them to integrate smoothly with the existing structure without disturbing the already-learned representations.


This selective training process helps the full model recover accuracy while maintaining the benefits of the smaller, efficient core. It also prepares the model to operate flexibly in real-world environments where it may need to run in smaller or larger configurations depending on the available compute resources.

In both fine-tuning stages, we ensure consistency in training setup by using the same optimizer, loss function, learning rate scheduling, and early stopping mechanism. This allows us to fairly compare the performance of pruned and rebuilt models across various experiments.

\section{Iterative Pruning and Rebuilding}

While one-shot pruning (i.e., removing a large portion of the network at once) can significantly reduce model size, it often leads to a sharp drop in accuracy that is difficult to recover. To address this limitation, we also adopt an \textbf{iterative pruning strategy}, where the model is pruned gradually over multiple steps. Each step involves pruning a small percentage of channels followed by fine-tuning for the final core model.

Iterative pruning gives better control over the balance between model size and accuracy. By pruning the network gradually, rather than all at once, the model can adapt step-by-step during fine-tuning. This gradual process helps maintain accuracy better because the model has time to adjust after each pruning step. Additionally, this method also creates smaller subnetworks within the original model, which supports adaptive use in different situations.

\subsubsection*{Overview of the Iterative Process}

\begin{figure}[ht]
    \centering
    \includegraphics[width=0.9\linewidth]{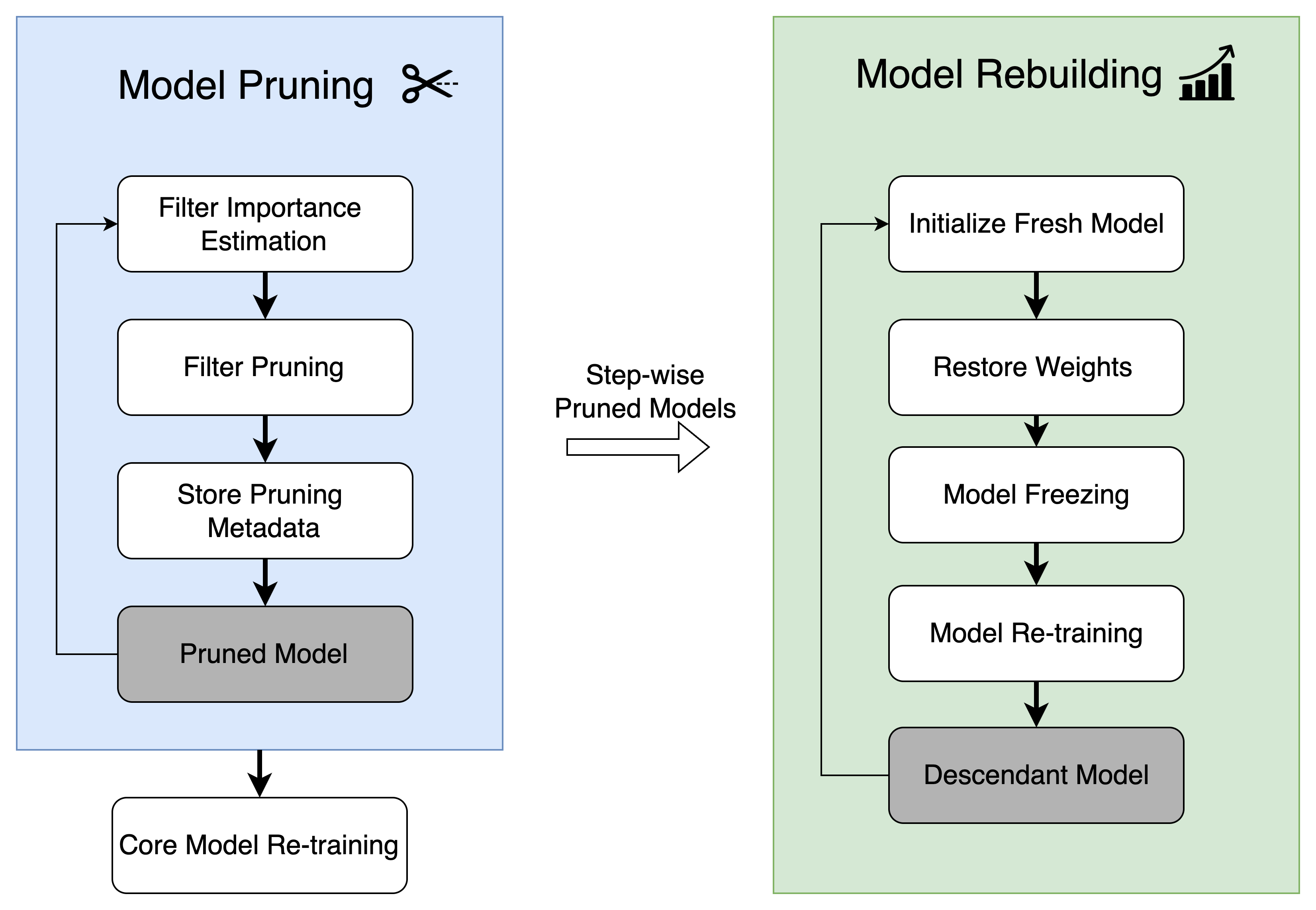}
    \caption{Overview of the iterative pruning and rebuilding process. The model is progressively compressed via structured pruning steps and later reconstructed in reverse using preserved weight metadata.}
    \label{fig:iterative_flow}
\end{figure}

The iterative pruning procedure begins with a fully pre-trained model. In each iteration:

\begin{itemize}
    \item A fixed percentage of filters, 20\% in our case are removed using our high level pruning method based on magnitude importance.
    \item The pruned model's accuracy, parameter count, and memory usage are recorded after each pruning step.
    \item The resulting smaller model serves as the starting point for the next pruning step.
    \item Information about the pruned and unpruned channels, including their weights and indices, is stored for later reconstruction.
\end{itemize}

After completing all pruning steps, the final pruned core model is fine-tuned for several epochs. This approach allows the model to adjust and recover performance effectively, ensuring better accuracy and robustness compared to one-shot pruning.

\subsubsection*{Rebuilding Procedure}

Once the final (most compressed) model is obtained, we apply our rebuilding procedure.

Starting from the smallest pruned model, we gradually reintroduce the previously removed filters using the stored metadata. At each level:
\begin{itemize}
    \item The number of filters is extended to match the earlier pruning state.
    \item The original weights: both pruned and retained are restored into the appropriate positions.
    \item The reconstructed model is fine-tuned to adjust to the reintroduced parameters and connections to improve accuracy.
\end{itemize}

This rebuilding is repeated in reverse order across all pruning steps, progressively reconstructing a version of the original full model. The resulting network is now composed of modular, scalable components that can be selectively activated or frozen based on runtime requirements.

This iterative pruning and rebuilding strategy not only ensures smooth compression with minimal accuracy loss, but also enables flexible and adaptive model deployment. By embedding multiple model scales within a single architecture, this approach lays the foundation for efficient inference on a wide range of hardware platforms: from constrained edge devices to high-performance servers.

Figure~\ref{fig:iterative_flow} illustrates the overall pipeline for our iterative pruning and rebuilding strategy. On the left, filters are pruned gradually using importance-based metrics, followed by fine-tuning at the end when we reach our final core model. On the right, the network is rebuilt in reverse by reintroducing the removed filters and weights at each pruning stage.

%% file: sections/experiments.tex
\chapter{Experimental Setup}\label{ch:experiments}

This chapter presents the experimental setup designed to evaluate pruning, rebuilding, and fine-tuning strategies with a focus on achieving model adaptivity. It outlines the models used (ResNet20, ResNet56, VGG-16, and AlexNet), the dataset (CIFAR-10), and the range of pruning strategies and importance metrics explored. It describes the dataset and data preprocessing steps, the hardware and software environment, pruning configurations, fine-tuning settings, and model evaluation protocols. It also details the hyperparameters chosen for training and fine-tuning, the different pruning percentages applied in both one-shot and iterative settings, and the evaluation metrics used to measure performance. Finally, we outline the evaluation protocols and metrics used to assess not only accuracy but also the effectiveness of adaptivity and how successfully they scale back toward their original size without significant loss in accuracy or stability.

\section{Dataset and Preprocessing}

All experiments in this study are conducted on the CIFAR-10 dataset, a widely used benchmark for image classification. The dataset consists of 60{,}000 color images, each of size 32$\times$32 pixels, categorized into 10 classes such as airplane, automobile, bird, cat, and so on. It is divided into 50{,}000 training images and 10{,}000 test images. For our experiments, the training set is further split into a training and validation set using an 80-20 ratio.

\subsection*{Data Augmentation}
To improve the generalization ability of the models and prevent overfitting, data augmentation techniques are applied to the training images. These include:
\begin{itemize}
    \item \textbf{Random horizontal flip} with a 50\% probability,
    \item \textbf{Random rotation} up to 15 degrees,
    \item \textbf{Color jittering} to slightly vary brightness, contrast, saturation, and hue.
\end{itemize}

\subsection*{Image Resizing}
Although CIFAR-10 images are originally 32$\times$32 pixels, some models such as AlexNet and ResNet20 expect inputs of size 224$\times$224. Therefore, the images are resized accordingly for these models. For VGG-16, which supports 32$\times$32 input resolution, resizing is skipped.

\subsection*{Normalization}
All input images are normalized using the standard ImageNet normalization statistics:
\begin{itemize}
    \item Mean: [0.485, 0.456, 0.406]
    \item Standard Deviation: [0.229, 0.224, 0.225]
\end{itemize}
This ensures that the input values are on a similar scale to those the pre-trained models were originally trained on.

\subsection*{Dataloader Configuration}
The dataset is loaded using PyTorch's \texttt{DataLoader}, with a batch size of 32 and 4 parallel workers for efficient data loading. The data loaders are defined separately for training, validation, and testing to ensure proper evaluation and reproducibility.

\section{Models and Architectures}

In this study, we experimented with three popular convolutional neural networks (CNNs): VGG-16, AlexNet, ResNet-20 and ResNet-56. These models were chosen because each has a unique architecture, varying complexity, and different amounts of parameters, allowing us to evaluate across diverse network structures. All models were trained and tested using the CIFAR-10 dataset. In this thesis, we use the VGG-16, ResNet-20 and ResNet-56 model pretrained on CIFAR-10, sourced from the \texttt{chenyaofo/pytorch-cifar-models} repository.

\subsection{VGG-16}
VGG-16 is a deep convolutional neural network known for its simplicity and structured design. It consists of 13 convolutional layers organized into five blocks. Each block includes multiple convolutional layers followed by a max-pooling layer. After these feature extraction blocks, the model uses three fully connected layers to perform classification.

We use the batch-normalized version of VGG-16, pretrained on CIFAR-10, as provided by the PyTorch Hub. During pruning, the convolutional layers are the main focus, while the classifier layers are kept unchanged to preserve the model's output format. When rebuilding the model, the classifier’s input dimensions are recalculated based on the reduced feature map sizes.

\subsection{ResNet-20}
ResNet-20 is a type of residual neural network known for its use of shortcut (or skip) connections. These connections help the model better handle deeper architectures by allowing gradients to flow more smoothly, thus addressing the vanishing gradient problem. As a result, ResNet models perform effectively even with fewer parameters compared to other deep models.

Specifically designed for smaller datasets like CIFAR-10, ResNet-20 is structured into three main stages, each composed of multiple residual blocks. Every block contains two convolutional layers along with an identity shortcut connection. In this thesis, we use the ResNet-20 model pretrained on CIFAR-10, available through PyTorch Hub.

Pruning ResNet-20 requires special attention due to its shortcut connections. These connections must remain properly aligned even after pruning, which means channel dimensions need careful handling. Our pruning and rebuilding process ensures that the skip connections remain consistent and fully functional, even after the network's structure has been reduced and subsequently reconstructed.

\subsection{ResNet-56}

ResNet-56 is a deeper variant of the ResNet family that also leverages residual connections to facilitate gradient flow through the network. Like ResNet-20, it is designed with shortcut (skip) connections that help preserve information across layers and mitigate issues such as the vanishing gradient problem.

ResNet-56 follows a similar architectural design as ResNet-20 but with increased depth. It consists of three main stages, each containing 9 residual blocks per stage compared to the 3 in ResNet-20. Each residual block includes two convolutional layers and an identity shortcut. This increased depth enables ResNet-56 to capture more abstract features, making it suitable for higher-capacity learning on datasets like CIFAR-10.

\subsection{AlexNet}

AlexNet is one of the early convolutional neural networks that demonstrated the effectiveness of deep learning for large-scale image classification. It consists of five convolutional layers followed by three fully connected layers. Due to its relatively shallow depth compared to modern architectures, it serves as a strong baseline model for evaluating pruning strategies at moderate scale.

Since there wasn't an existing version of AlexNet pretrained on the CIFAR-10 dataset, we trained our own version specifically for this task. We adopted the version of AlexNet pretrained on ImageNet and adapted it to CIFAR-10 by fine-tuning all layers. The input images were resized to 224×224 pixels and normalized using ImageNet mean and standard deviation to match the pretrained weight requirements.

Training was performed using PyTorch Lightning with early stopping and model checkpointing based on validation loss. The final model served as the starting point for all pruning, rebuilding, and fine-tuning experiments involving AlexNet.

This approach allowed us to create a strong baseline, enabling a clear comparison of how effective our pruning and adaptive rebuilding methods were when applied to medium-sized neural networks.

\section{Hardware and Software Configuration}

All the experiments in this research were conducted on the Snellius supercomputer provided by SURF \cite{surf2025snellius}, using nodes with NVIDIA A100 GPUs. These powerful GPUs allowed for fast training and testing of deep learning models, as well as efficient pruning and rebuilding.

For building and training neural networks, we used PyTorch \cite{paszke2019pytorchimperativestylehighperformance}. To make the training process easier and more organized, we used PyTorch Lightning \cite{Falcon_PyTorch_Lightning_2019}, which helped with tasks like early stopping, saving the best models, and logging progress. For pruning the models, we used the Torch-Pruning library \cite{depGraph}, which supports different types of structured pruning methods.

Weights \& Biases (WandB) was used to track the results of training and pruning, and to log useful metrics like accuracy and loss. We used the Torchvision library to load standard datasets like CIFAR-10 and to download pre-trained models.

\section{Training and Fine-Tuning Configuration}

To train and fine-tune our models, we used standard supervised learning with cross-entropy loss as the objective function. The models were optimized using the Adam optimizer, which provides good convergence with minimal tuning. We experimented with multiple learning rates, including $1e{-2}$, $1e{-3}$, $1e{-4}$, and $1e{-5}$ depending on the model and pruning stage.

We also explored various schedulers for adjusting the learning rate during training, including cosine annealing, step decay, exponential decay, and cyclic schedules. After experimentation, we found that cosine annealing consistently gave the best results and therefore we used it for most of our final experiments.

The fine-tuning process involved two main stages. First, we fine-tuned immediately after pruning to help the model recover performance following the removal of certain parameters. Next, after rebuilding the model, we fine-tuned again to allow the reintroduced channels to smoothly integrate with the previously learned parameters.

To prevent overfitting and unnecessary training, we implemented early stopping to stop training once validation loss no longer improved. Throughout training and fine-tuning, we used PyTorch Lightning to simplify the process and logged important metrics like loss, accuracy, and learning rate using Weights and Biases (WandB) \cite{wandb}.

By using a consistent training setup for all models, we ensured fair comparisons of the effectiveness of different pruning strategies and fine-tuning approaches.

\section{Evaluation Metrics}

To evaluate the adaptivity of our models under various pruning strategies, we employed a set of metrics that reflect both model efficiency and performance. The most important one was test accuracy, which measures how well the model generalizes to unseen data. This allowed us to assess whether the pruned and subsequently rebuilt models retained their generalization capabilities. In addition to accuracy, we also tracked the number of parameters in the model. This gave us an idea of how much the model had been compressed, since fewer parameters usually mean a smaller and faster model.

We also measured the model size in megabytes (MB), which records how much memory the model takes up when saved. This is particularly relevant in resource-constrained environments such as edge devices, where adaptivity must consider memory limitations as well as computational ones. During training and fine-tuning, we recorded training and validation losses to check how well the model was learning and to monitor signs of overfitting. We also tracked training and validation accuracies to see how the model was performing during each epoch.

All of these metrics were logged using Weights \& Biases (WandB), which allowed us to visualize and compare the performance of different pruning strategies and model configurations in a clear and organized way.

%% file: sections/evaluation.tex
\chapter{Evaluation}\label{ch:evaluation}

\section{Overview of Evaluation Goals}

The evaluation in this thesis focuses on assessing the effectiveness of structured pruning and the ability to rebuild pruned neural networks without significant loss in performance. The central question is: can we reduce a model's size using pruning techniques, and later reconstruct it back to its original shape while retaining or even improving accuracy?

Our experiments are structured to evaluate three key stages:

\begin{itemize}
    \item \textbf{Pruning:} We apply different structured pruning techniques including manual, high-level importance-based, and soft pruning across various compression levels (30\%, 50\%, 70\%). We analyze how much each method reduces the number of parameters and model size, and how accuracy is impacted.

    \item \textbf{Rebuilding:} Using stored pruning metadata, we reconstruct the original architecture by reintroducing the pruned weights. We then fine-tune the rebuilt model and evaluate whether it can match or exceed the performance of the pruned model or the original baseline.

    \item \textbf{Iterative Pruning:} Does gradually pruning the network in smaller steps help the model maintain better accuracy as compared to one-shot? We investigate this to verify whether our method supports multiple modes of operation in real-world applications where resource availability may vary dynamically.
\end{itemize}

The evaluation does not explicitly test adaptivity in dynamic runtime environments. However, by showing that rebuilt models preserve the pruned sub-network internally, our results demonstrate the potential for flexible and adaptive deployment in future work.

Throughout this evaluation, we measure performance using standard metrics such as test accuracy, number of parameters, and model size in megabytes (MB). The experiments are performed consistently across architectures like VGG-16, ResNet-20, ResNet-56 and AlexNet to ensure fair comparisons.

\section{Baseline Model Performance}

Before applying any pruning or reconstruction techniques, we first evaluated the performance of the original, pre-trained models on the CIFAR-10 dataset.

We tested four different architectures: VGG-16, AlexNet, ResNet-20 and ResNet-56. ResNet-20 and VGG-16 were initialized with pre-trained weights. However, since no CIFAR-10 variant of AlexNet was publicly available, we fine-tuned an ImageNet-pretrained version on CIFAR-10 ourselves.

The models were evaluated using standard classification metrics. For each model, we measured test accuracy, the total number of trainable parameters, and the model size in megabytes (MB). These metrics are summarized in Table~\ref{tab:baseline-performance}.

\vspace{0.5em}
\begin{table}[ht]
\centering
\caption{Baseline performance of original (unpruned) models on CIFAR-10}
\label{tab:baseline-performance}
\begin{tabular}{|c|c|c|c|}
\hline
\textbf{Model} & \textbf{Test Accuracy (\%)} & \textbf{Parameters (M)} & \textbf{Model Size (MB)} \\
\hline
VGG-16         & 93.94 & 15.25 & 58.244 \\
AlexNet        & 90.16 & 57.04 & 217.614 \\
ResNet-20      & 92.60 & 0.27 & 1.078 \\
ResNet-56      & 86.49 & 0.86 & 3.369 \\
\hline
\end{tabular}
\end{table}
\vspace{0.5em}

These baseline results help us understand how much accuracy is changed after pruning, and how effective our fine-tuning strategies are in recovering lost performance. All later comparisons are made relative to these baseline values.

\section{One-Shot Pruning Results}
This section presents the results from applying one-shot pruning, where a large portion of parameters was pruned in a single step. We evaluate the effectiveness of this approach by looking at model accuracy, parameter reduction, and model size.

\subsection{Accuracy vs. Pruning Percentage}

\begin{figure}[h!] \centering \includegraphics[width=0.9\textwidth]{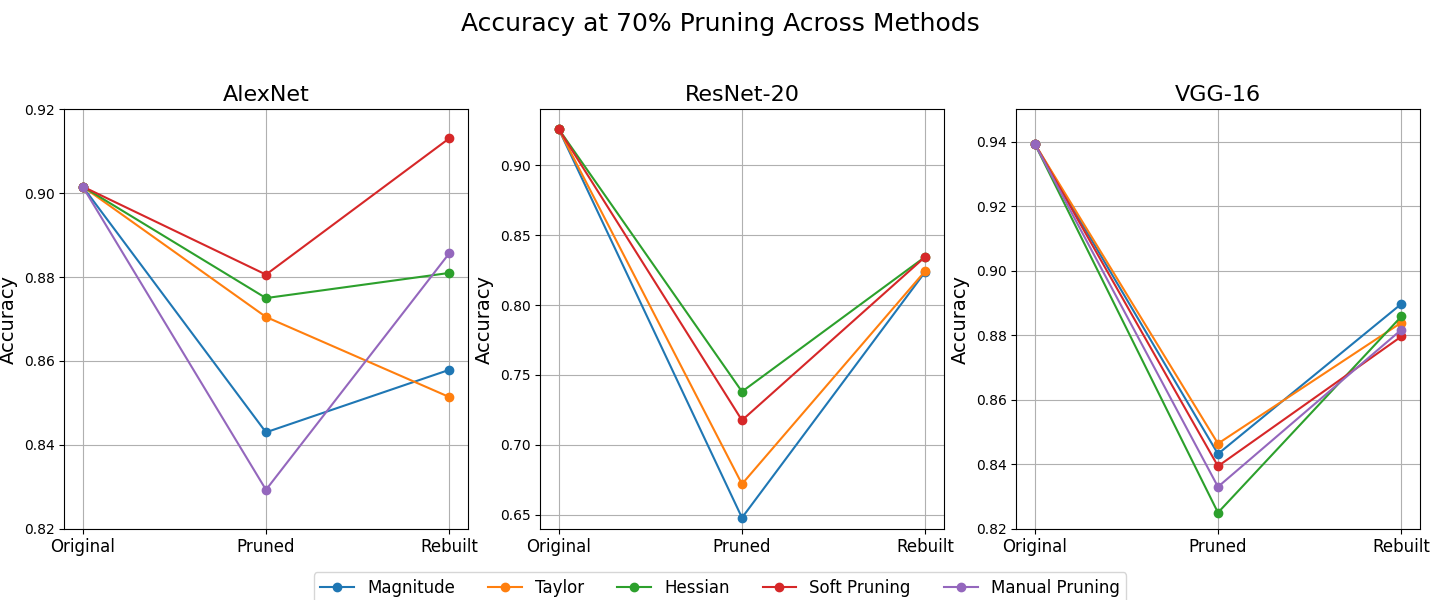}\caption{Accuracy progression across different pruning strategies on three architectures (AlexNet, ResNet-20, and VGG-16) at 70\% pruning. Each method shows three accuracy points: the original (unpruned) model, accuracy immediately after pruning, and accuracy after rebuilding and fine-tuning. Results highlight how structured pruning combined with rebuilding enables recovery of model performance} \label{fig:accuracy-pruning} \end{figure}

The accuracy of each model was measured on the CIFAR-10 test set before and after one-shot pruning. Figure~\ref{fig:accuracy-pruning} shows the graphs for 70\% pruning across different pruning methods used for all the models. Table~\ref{tab:prune-vs-rebuild} shows how accuracy changes as we increase the pruning percentage from 30\% to 70\%. Each curve represents a pruning method's performance from the original model to its pruned and rebuilt versions.

We observe that all methods consistently recover and even surpass the pruned accuracy after rebuilding. Manual pruning, while effective for VGG, is not applicable to all architectures due to structural constraints. ResNet-20 shows a comparatively higher drop during pruning, but also demonstrates effective recovery with appropriate fine-tuning.

These trends emphasize the importance of combining structured pruning with a careful rebuilding pipeline to regain performance after aggressive compression.

\begin{table}[h!]
\centering
\caption{Test Accuracy (\%) Before and After Rebuilding for Each Model and Method}
\label{tab:prune-vs-rebuild}
\begin{tabular}{|l|c|c|c|c|c|c|c|c|}
\hline
\multirow{2}{*}{\textbf{Method}} & \multirow{2}{*}{\makecell{\textbf{Pruning} \\ \textbf{(\%)}}} 
& \multicolumn{2}{c|}{\textbf{VGG-16}} 
& \multicolumn{2}{c|}{\textbf{AlexNet}} 
& \multicolumn{2}{c|}{\textbf{ResNet-20}} \\
\cline{3-8}
 & & Pruned & Rebuilt & Pruned & Rebuilt & Pruned & Rebuilt \\
\hline
\textbf{\makecell{\textbf{Baseline} \\ \textbf{(0\%)}}} & 0 & 93.94 & -- & 90.16 & -- & 91.5 & -- \\
\hline
\multirow{2}{*}{\makecell{\textbf{High Level:} \\ \textbf{Magnitude}} } 
& 30 & 88.56 & 88.8 & 90.72 & 85.59 & 80.25 & 82.28 \\
& 50 & 87.46 & 89.04 & 89.76 & 84.58 & 77.46 & 81.93 \\
& 70 & 84.32 & 88.97 & 84.3 & 85.79 & 64.79 & 82.39 \\
\hline
\multirow{2}{*}{\makecell{\textbf{High Level:} \\ \textbf{Taylor}} } 
& 30 & 89.85 & 90.27 & 90.49 & 84.34 & 81.99 & 82.89, \\
& 50 & 87.59 & 88.44 & 89.18 & 84.45 & 76.48 & 82.65 \\
& 70 & 84.64 & 88.39 & 87.05 & 85.14 & 67.2 & 82.45 \\
\hline
\multirow{2}{*}{\makecell{\textbf{High Level:} \\ \textbf{Hessian}} } 
& 30 & 89.63 & 88.34 & 87.8 & 88.3 & 82.86 & 83.26 \\
& 50 & 88.35 & 87.78 & 85.9 & 86.7 & 73.03 & 82.58 \\
& 70 & 82.5 & 88.58 & 87.5 & 88.1 & 73.81 & 83.46 \\
\hline
\multirow{2}{*}{\makecell{\textbf{Manual} \\ \textbf{Pruning}} } 
& 30 & 86.5 & 90.1 & 91.35 & 92.3 & -- & -- \\
& 50 & 85.4 & 89.12 & 88.84 & 90.12 & -- & -- \\
& 70 & 83.3 & 88.17 & 88.06 & 91.31 & -- & -- \\
\hline
\multirow{2}{*}{\makecell{\textbf{Soft} \\ \textbf{Pruning}} } 
& 30 & 87.81 & 78.07 & 87.36 & 92.78 & 72.97 & 82.75 \\
& 50 & 88.08 & 88.84 & 85.71 & 91.98 & 71.26 & 82.83 \\
& 70 & 83.94 & 87.96 & 82.93 & 88.57 & 71.76 & 83.46 \\
\hline
\end{tabular}
\end{table}

As expected, accuracy decreases as more parameters are pruned. AlexNet and VGG-16 shows the highest resilience, maintaining a relatively higher accuracy even at high pruning percentages, while ResNet-20 show sharper accuracy drops.

\textit{Note:} Manual pruning was not performed on ResNet-20. While the pruning framework (DepGraph) successfully identified and pruned layers in ResNet-20, the returned pruning indices were inconsistent across iterations. This inconsistency stems from the model’s highly interconnected structure and the use of in-place pruning within a loop, which caused the reference indexing to shift after each pruning step. As a result, accurate reconstruction of the pruned model was not feasible for this architecture using manual pruning.

\subsection{Parameter Reduction and Size}

The goal of pruning is not only to maintain accuracy but also to significantly reduce model size and parameters. Table~\ref{tab:params-size-pruning} summarizes how parameter count and model size are reduced after pruning.

\begin{table}[h!]
\centering
\caption{Parameter Count and Model Size After One-Shot Pruning}
\label{tab:params-size-pruning}
\begin{tabular}{|c|c|c|c|c|c|}
\hline
\textbf{Model} & \makecell{\textbf{Pruning} \\ \textbf{(\%)}} & \makecell{\textbf{Original} \\ \textbf{Parameter} \\ \textbf{Count (M)}} & \makecell{\textbf{Parameters} \\ \textbf{After Pruning}} & \makecell{\textbf{Original} \\ \textbf{Size (MB)}} & \makecell{\textbf{Size After} \\ \textbf{Pruning}} \\
\hline
\multirow{4}{*}{VGG-16} 
    & 30 & 15.25 & 7.71  & 58.24  & 29.45 \\
    & 50 &          & 4.09  &        & 15.65 \\
    & 70 &          & 1.69  &        & 6.49 \\
\hline
\multirow{4}{*}{AlexNet} 
    & 30 & 57.04 & 44.66 & 217.61 & 170.38 \\
    & 50 &          & 36.41 &        & 138.90 \\
    & 70 &          & 28.49 &        & 108.69 \\
\hline
\multirow{4}{*}{ResNet-20} 
    & 30 & 0.27   & 0.12    & 1.07 & 0.53 \\
    & 50 &          & 0.063     &      & 0.28 \\
    & 70 &          & 0.023     &      & 0.12 \\
\hline
\end{tabular}
\end{table}

\begin{figure}[h!] \centering \includegraphics[width=0.9\textwidth]{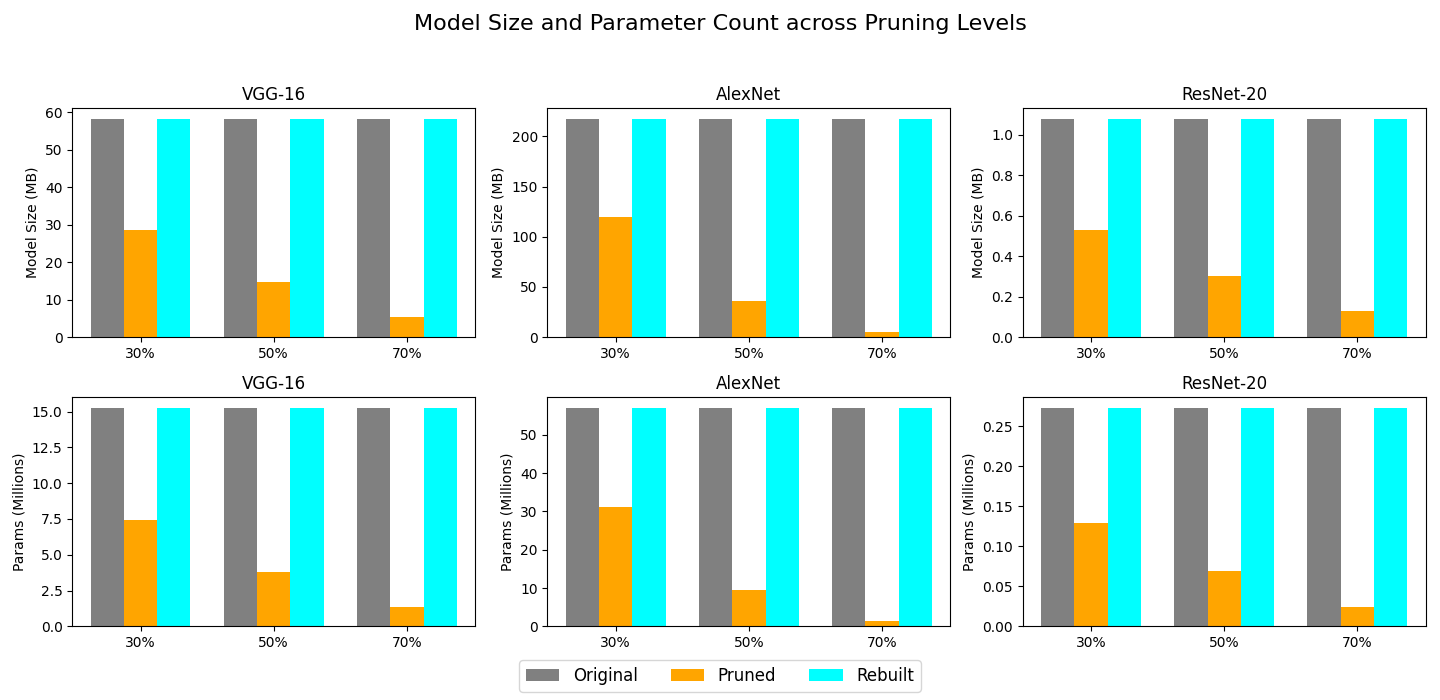}
\caption{Comparison of parameter count and model size across different pruning stages (30\%, 50\%, 70\%) for VGG-16, AlexNet, and ResNet-20.  The plot demonstrates how model size is reduced through pruning and restored via rebuilding. Pruning reduces model complexity, while rebuilding restores original structure. Top graph: Number of parameters before pruning (gray), after pruning and fine-tuning (orange), and after rebuilding and fine-tuning (blue) at different pruning percentages for all models. Bottom: Model size (in MB) across the same pruning stages. Rebuilt models return to original size. Bars are grouped by model from left to right: \textbf{VGG-16}, \textbf{AlexNet}, and \textbf{ResNet-20}.} \label{fig:params-pruning} \end{figure}

This table clearly illustrates the effectiveness of pruning in reducing both the number of parameters and the physical memory size of the models. For example, pruning 70\% of parameters from AlexNet reduces its size from 217.61 MB down to 108.69 MB, showing significant potential for efficiency improvements.

It's important to note that although we applied the same pruning percentages (30\%, 50\%, and 70\%) across different methods, the final number of parameters and model sizes varied slightly. This is because each pruning method decides which parameters to remove based on different ways of measuring their importance. Some methods might prune more channels from earlier layers, while others might focus on later layers, resulting in slight differences in the final model size and parameter count. The table shown above presents the numbers from Manual Pruning. However, while other methods gave slightly different numbers, the overall sizes and parameter counts remained similar.

Figure~\ref{fig:params-pruning} presents two bar plots. The top plot shows how the number of parameters changes from the original model, to the pruned version, and finally to the rebuilt model across multiple pruning percentages. The bottom plot provides a similar comparison in terms of model size (in MB).

As shown in the graphs, all three models experience a significant drop in parameters and size after pruning. Rebuilding increases the parameter count and memory footprint back to the original level, since the pruned channels are reinserted. Interestingly, despite the model size returning to its original value after rebuilding, the rebuilt models still retain much of the efficiency learned during pruning and fine-tuning. This validates the effectiveness of our proposed adaptive pruning and reconstruction framework.

\section{Iterative Pruning and Rebuilding}

This section presents the results and insights from our iterative pruning experiments, along with the rebuilding process to recover and adapt the network to larger configurations. As introduced in the methodology, iterative pruning gradually reduces the model’s size by removing a small percentage of channels in each step, rather than pruning large portions all at once. This approach helps the model retain performance by avoiding sudden drops in capacity. More importantly, iterative pruning enables support for multiple modes of operation, which is crucial for real-world deployment.

\subsection*{Pruning in Stages}

In our experiments, we performed structured pruning in multiple iterations to explore whether our approach can generalize across multiple modes of adaptivity. Instead of focusing solely on recoverable accuracy, our goal was also to evaluate if iterative pruning can support flexible scaling of the model to meet varying resource and performance requirements.

We tracked accuracy, number of parameters, and model size after each step. We observed that:
\begin{itemize}
    \item Accuracy decreased more gracefully across steps compared to one-shot pruning.
    \item The final core model had higher accuracy at the same pruning level compared to one-shot pruning.
\end{itemize}

\subsection*{Rebuilding and Adaptivity}

After the core model was trained, we performed the rebuilding process to get the model back to larger configurations using the pruning metadata stored at each step. At each level of rebuilding, we reinserted previously removed channels and restored their weights, followed by a short round of fine-tuning.

This process allowed us to test how well the model can “grow back” and recover performance when more resources are available. Our results showed that:
\begin{itemize}
    \item Rebuilt models consistently achieved higher accuracy than their corresponding pruned counterparts.
    \item Fine-tuning after rebuilding helped the new channels integrate smoothly with the already-learned weights.
\end{itemize}

To evaluate the impact of iterative pruning and rebuilding, we applied it to all 4 models. The models was pruned in three gradual steps with 20\% pruning in each step and fine tuning after each step to regain accuracy. Then, using the saved pruning metadata, we reversed the process by incrementally restoring previously pruned channels, followed by fine-tuning at each rebuild stage.

During the rebuilding process, we selectively froze the core weights. Freezing these channels helped maintain adaptability and allowing the model to scale seamlessly across configurations.

\begin{table}[h!]
\centering
\caption{Results from Iterative Pruning and Rebuilding on ResNet-20}
\label{tab:iterative-results_ResNet56}
\begin{tabular}{|c|c|c|c|}
\hline
\textbf{Stage} & \textbf{Accuracy (\%)} & \textbf{Parameters (M)} & \textbf{Model Size (MB)} \\
\hline
Original Model   & 86.50 & 0.86  & 3.37 \\
Forward Step 3   & 71.80 & 0.21  & 0.91 \\
Backward Step 3  & 75.77 & 0.33  & 1.37 \\
Backward Step 2  & 78.65 & 0.54  & 2.16 \\
Backward Step 1  & 80.13 & 0.86  & 3.37 \\
\hline
\end{tabular}
\end{table}

\begin{figure}[h!]
    \centering
    \includegraphics[width=\textwidth]{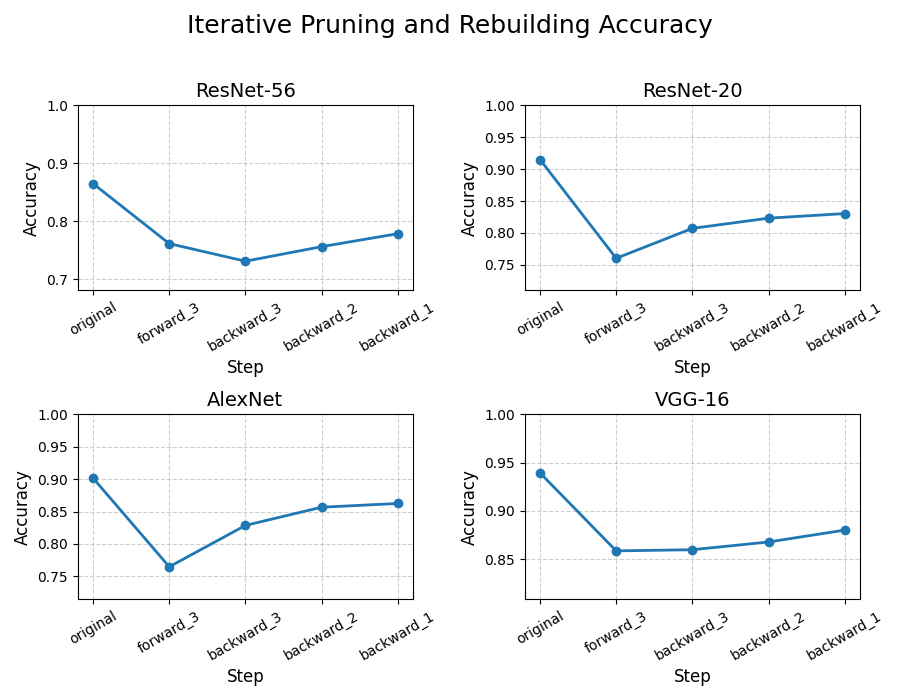}
    \caption{Accuracy progression for ResNet-20, ResNet-56, VGG-16, and AlexNet across iterative pruning and rebuilding stages.}
    \label{fig:iterative-accuracy}
\end{figure}

\begin{figure}[h!]
    \centering
    \includegraphics[width=\textwidth]{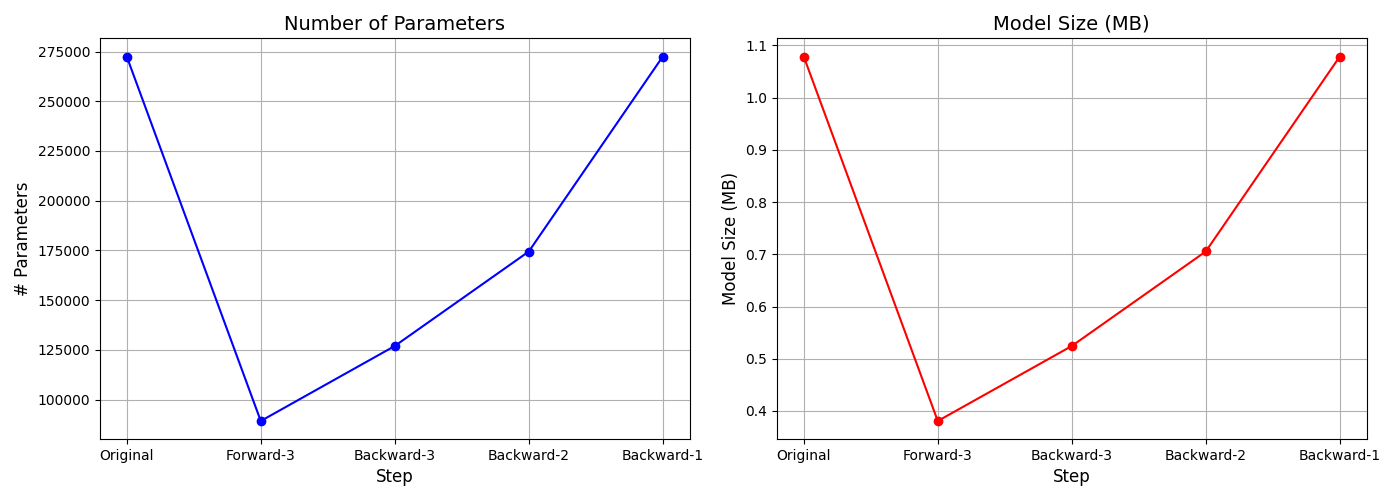} \\
    \vspace{0.5em}
    \caption{Profiling metrics of AlexNet showing changes in parameter count and model size across pruning levels.}
    \label{fig:iterative-params}
\end{figure}

Figure~\ref{fig:iterative-accuracy} and Figure~\ref{fig:iterative-params} illustrates the effectiveness of our rebuilding strategy. The graph Figure~\ref{fig:iterative-params} demonstrates how the model’s size and complexity (number of parameters and model size) increase step-by-step during rebuilding, starting from the smallest, most compressed core model (forward\_3). The graph Figure~\ref{fig:iterative-accuracy} shows the corresponding changes in accuracy at each rebuilding stage for different network architectures. Initially, accuracy is lowest at the highly pruned forward\_3 step, as expected. As we progressively rebuild the model by reintroducing previously pruned channels (steps backward\_3 to backward\_1), the accuracy improves significantly.

Table~\ref{tab:iterative-results_ResNet56} shows the changes in accuracy, number of parameters, and model size (in MB) throughout this process for ResNet-56. We observed that accuracy decreased gradually during pruning. Also, when we rebuilt the model, accuracy consistently improved again. This shows that our iterative pruning and rebuilding method allows the model to adapt flexibly, maintaining performance while adjusting to different sizes.

\section{Ablation Studies}

To explore how pruning decisions affect overall performance, we conducted ablation studies in which instead of pruning all convolutional layers, we experimented with a manual pruning setup where only alternate layers were pruned. This was done to test whether skipping some layers during pruning could help preserve important parts of the network, and in turn, lead to better accuracy after pruning.

In standard structured pruning, every convolutional layer is pruned, regardless of its position or influence in the network. The early layers usually capture low-level features like edges or textures that are reused throughout the network, while the deeper layers are more focused on the specific task. So if we prune too many filters from all layers equally, we might end up removing useful information that the model still needs.

We tested this on three models: AlexNet, ResNet-20, and VGG-16. The pruning percentages remained the same (30\%, 50\%, 70\%), but pruning was skipped for alternate layers. All other settings, such as learning rate, scheduler, and fine-tuning duration, were kept consistent with the earlier experiments. As usual, each model was fine-tuned after pruning, and then rebuilt to its original size using the saved pruning metadata. The results and comparison of accuracy with the Magnitude based High Level Pruner are shown in Table~\ref{tab:alternate-layer-results} and comparison of parameters and model size is shown in Table~\ref{tab:ablation-vs-original}

\begin{table}[h!]
\centering
\caption{Test Accuracy (\%) Comparison for Alternate Layer Pruning on VGG-16, AlexNet, and ResNet-20 vs Magnitude Based High Level Pruner}
\label{tab:alternate-layer-results}
\begin{tabular}{|c|c|c|c|c|c|c|}
\hline
\textbf{Model} & \textbf{\makecell{\textbf{Pruning} \\ \textbf{(\%)}}} & \textbf{\makecell{\textbf{Original} \\ \textbf{Accuracy}}} & \textbf{\makecell{\textbf{Pruned} \\ \textbf{Accuracy}}} & \textbf{\makecell{\textbf{Rebuilt} \\ \textbf{Accuracy}}} &
\textbf{\makecell{\textbf{Pruned} \\ \textbf{Accuracy} \\ \textbf{(Before)}}} &
\textbf{\makecell{\textbf{Rebuild} \\ \textbf{Accuracy} \\ \textbf{(Before)}}} \\
\hline
\multirow{3}{*}{AlexNet} 
& 30 & \multirow{3}{*}{90.16} & 90.58 & 96.64 & 90.72  & 85.59\\
& 50 &                        & 90.29 & 95.18 & 89.76 & 84.58 \\
& 70 &                        & 89.48 & 95.08 & 84.3 & 85.79\\
\hline
\multirow{3}{*}{VGG-16} 
& 30 & \multirow{3}{*}{93.94} & 89.24 & 89.76 & 88.56 & 88.8 \\
& 50 &                        & 85.44 & 88.57 & 87.46 & 89.04\\
& 70 &                        & 82.76 & 88.38 & 84.32 & 88.97\\
\hline
\multirow{3}{*}{ResNet-20} 
& 30 & \multirow{3}{*}{92.60} & 83.12 & 84.12 & 80.25 & 82.28 \\
& 50 &                        & 75.35 & 80.77 & 77.46 & 81.93\\
& 70 &                        & 73.14 & 78.18 & 64.79 & 82.39\\
\hline
\end{tabular}
\end{table}

The results of this selective pruning strategy are summarized below. Surprisingly, this simple modification led to improved accuracy after pruning in most cases when compared to standard uniform pruning at the same percentages. Additionally, the rebuilt models were able to achieve or even exceed baseline accuracy in some cases, showing the strength of this more careful pruning design. On AlexNet, for example, pruning alternate layers led to better pruned accuracy compared to many high-level pruning methods. More notably, after rebuilding, AlexNet even surpassed its original baseline accuracy, reaching up to 96.64\% at 30\% pruning. VGG-16 also showed promising results. Even after pruning 70\% of its parameters from alternate layers, the rebuilt model still reached over 88\% accuracy, recovering much of the lost performance. For ResNet-20, the gains were more modest, but the model still demonstrated stable recovery after rebuilding. While pruned performance dipped at higher pruning levels, rebuilding helped regain much of the lost accuracy.

\begin{table}[h!]
\centering
\caption{Comparison of One-Shot vs Ablation Pruning Results}
\label{tab:ablation-vs-original}
\begin{tabular}{|l|r|r|r|r|r|}
\hline
\textbf{Model} & \makecell{\textbf{Pruning} \\ \textbf{\%}}  & \makecell{\textbf{Params} \\ \textbf{Before (M)}} & \makecell{\textbf{Size} \\ \textbf{Before (M)}} & \makecell{\textbf{Params} \\ \textbf{Ablation (M)}} & \makecell{\textbf{Size} \\ \textbf{Ablation (MB)}} \\
\hline
VGG-16    & 0  & 15.25   & 58.24  & 15.25   & 58.24  \\
          & 30 & 7.71    & 29.45  & 7.46    & 28.50  \\
          & 50 & 4.09    & 15.65  & 3.82    & 14.61  \\
          & 70 & 1.69    & 6.49   & 1.36    & 5.24   \\
\hline
AlexNet   & 0  & 57.04   & 217.61 & 57.04   & 217.61 \\
          & 30 & 44.66   & 170.38 & 56.32   & 214.84 \\
          & 50 & 36.41   & 138.90 & 55.82   & 212.95 \\
          & 70 & 28.49   & 108.69 & 55.34   & 211.09 \\
\hline
ResNet-20 & 0  & 0.27    & 1.07   & 0.27    & 1.08   \\
          & 30 & 0.12    & 0.53   & 0.19    & 0.78   \\
          & 50 & 0.063   & 0.28   & 0.14    & 0.57   \\
          & 70 & 0.023   & 0.12   & 0.088   & 0.38   \\
\hline
\end{tabular}
\end{table}

\begin{figure}[h!] \centering \includegraphics[width=0.9\textwidth]{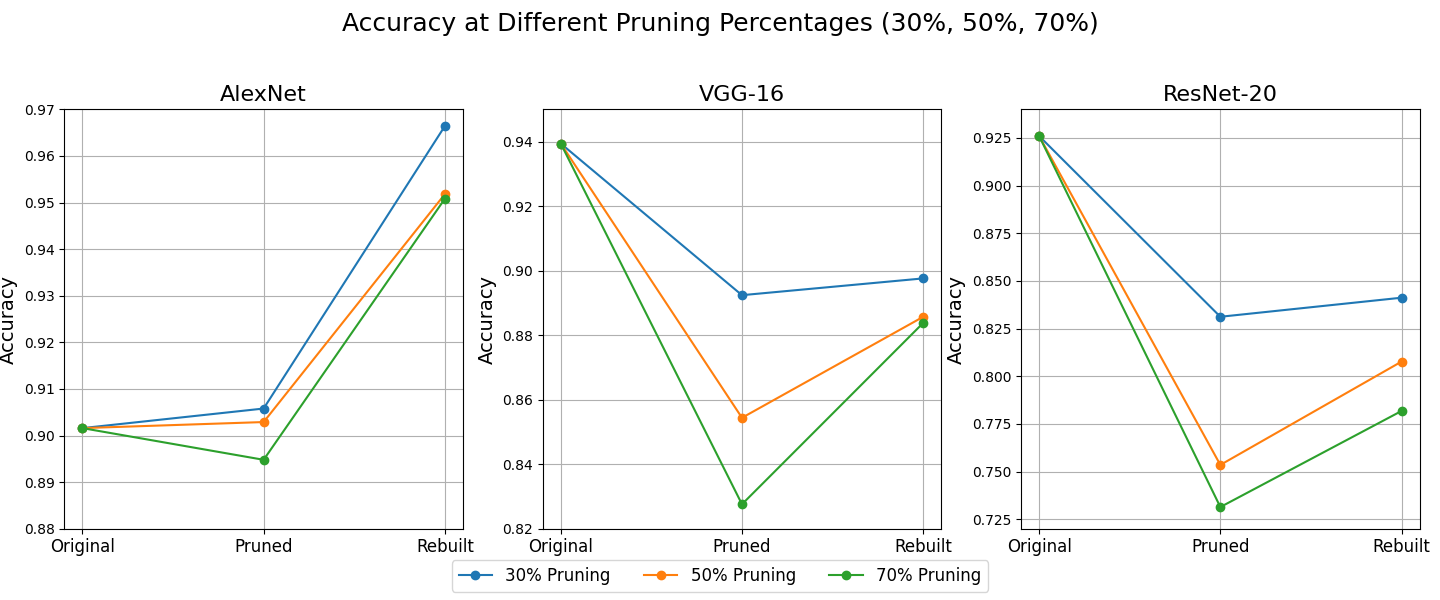}\caption{Ablation study on partial pruning: alternating layers are pruned while the rest are left intact. Accuracy is compared across the original, pruned, and rebuilt stages for AlexNet, VGG-16, and ResNet-20. Reconstruction continues to recover performance even when only a subset of layers is pruned.} \label{fig:ablation_study} \end{figure}

As shown in Figure \ref{fig:ablation_study}, all models experience a drop in accuracy after partial pruning, as expected. However, the drop is relatively less severe compared to uniform full-layer pruning, indicating that pruning fewer layers preserves more of the representational capacity of the original network.

Interestingly, the reconstruction step still leads to a noticeable recovery of performance across all models. For instance, AlexNet nearly regains its original accuracy, and both VGG-16 and ResNet-20 show marked improvement over the pruned baselines. These results suggest that even when pruning is applied selectively, our rebuilding approach remains effective and generalizable.

In addition to evaluating partial pruning, we also explored another ablation setting related to the iterative pruning pipeline. Typically, in our main experiments, fine-tuning was performed only once, after completing all pruning steps on the final core model. However, we conducted an experiment where instead of waiting until the end, we fine-tuned the model after each individual pruning step. The goal was to allow the network to gradually adapt at each stage, improving the quality of the intermediate core model, and ultimately helping the rebuilding phase start from a more accurate baseline. This experiment helped us assess whether finer-grained adaptation during pruning could lead to better accuracy recovery during reconstruction. Table~\ref{tab:iterative-results} shows the changes in accuracy, number of parameters, and model size (in MB) throughout this process for ResNet-20.

\begin{table}[h!]
\centering
\caption{Results from Iterative Pruning and Rebuilding on ResNet-20}
\label{tab:iterative-results}
\begin{tabular}{|c|c|c|c|}
\hline
\textbf{Stage} & \textbf{Accuracy (\%)} & \textbf{Parameters (M)} & \textbf{Model Size (MB)} \\
\hline
Original Model   & 91.5 & 0.27  & 1.08 \\
Forward Step 1   & 82.14 & 0.17  & 0.71 \\
Forward Step 2   & 79.91 & 0.12  & 0.52 \\
Forward Step 3   & 76.02 & 0.089   & 0.38 \\
Backward Step 3  & 80.71 & 0.12  & 0.52 \\
Backward Step 2  & 82.31 & 0.17  & 0.71 \\
Backward Step 1  & 83.03 & 0.27  & 1.08 \\
\hline

\end{tabular}
\end{table}
\begin{figure}[h!]
    \centering
    \includegraphics[width=0.75\linewidth]{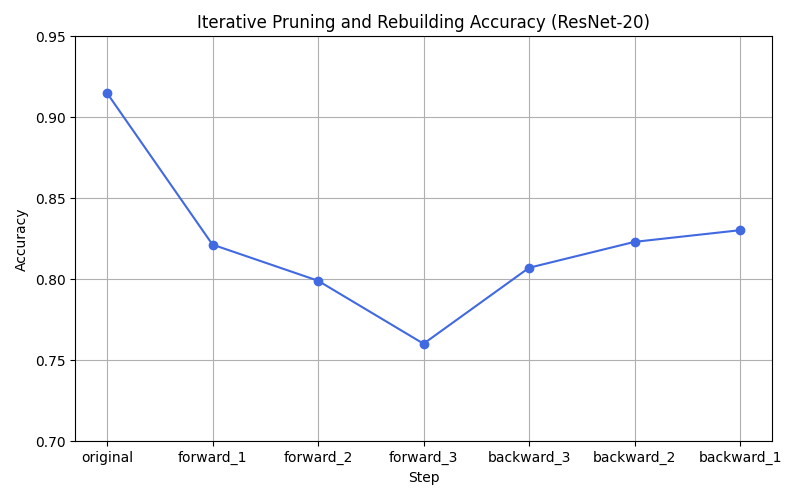}
    \caption{Profiling metrics of ResNet-20 showing changes in parameter count and model size across pruning levels.}
    \label{fig:iterative-res56}
\end{figure}

As shown in graph \ref{fig:iterative-res56}, this approach led to better accuracy recovery during reconstruction. By fine-tuning after each pruning step, the network had the opportunity to re-stabilize and adapt its weights to the reduced structure at every stage. This gradual adjustment helped prevent drastic accuracy drops and reduced the risk of losing critical information too early. As a result, the final pruned core preserved more meaningful representations, making the subsequent rebuilding phase more effective and less reliant on relearning from scratch. Compared to fine-tuning only once at the end, this finer-grained adaptation offered a smoother transition through model compression and ultimately yielded improved performance.

%% file: sections/discussion.tex
\chapter{Discussion}\label{ch:discussion}

This chapter reflects on the key findings from our structured pruning experiments using convolutional neural networks, examining how different pruning strategies such as high-level importance-based pruning, soft pruning, and iterative pruning impacted model accuracy and efficiency. It also explores how adaptive techniques helped recover performance, and how choices such as pruning alternate layers influenced outcomes. Furthermore, we discuss the practical challenges faced during pruning and fine-tuning, and consider the broader implications of adaptive pruning for deploying efficient deep learning models in real-world resource-constrained environments.

\section{Key Findings}

This section summarizes the main insights and trends observed during the evaluation phase of this thesis.

\begin{itemize}
    \item \textbf{All pruning methods reduce model complexity effectively.} As shown in Figure~\ref{fig:params-pruning}, pruning significantly reduced both the number of parameters and the model size across all architectures. For example, pruning 70\% of AlexNet reduced its size from 217~MB to 108~MB.

    \item \textbf{Rebuilding helps restore performance.} Rebuilding pruned models using saved metadata consistently improved their accuracy. In many cases, the rebuilt models came close to, or even surpassed, their original baseline performance, showing the benefit of allowing models to "grow back" after pruning.

    \item \textbf{Some methods work better than others.} Table~\ref{tab:prune-vs-rebuild} shows how different pruning methods affect accuracy. High-level pruning with Taylor or Hessian importance resulted in more stable accuracy post-rebuilding than simpler magnitude-based pruning.

    \item \textbf{Iterative pruning performs better than one-shot.} Gradual pruning in multiple steps helped the model retain more accuracy along the way (also discussed in \cite{lotteryTicket}. This method gave the network time to adjust between steps and consistently led to better accuracy recovery after rebuilding.

    \item \textbf{Alternate layer pruning was surprisingly effective.} In our ablation study, pruning only every other layer led to better results in some cases especially for VGG-16 and AlexNet. This suggests that skipping layers might reduce disruption in the network while still providing efficiency benefits.

    \item \textbf{Fine-tuning is critical.} The choice of learning rate, optimizer, and scheduler made a noticeable difference in how much performance could be recovered. Cosine learning rate scheduling worked best across nearly all experiments.

    \item \textbf{Each architecture behaves differently.} Some models handled pruning better than others. VGG-16 showed strong resilience, while ResNet-20, with its skip connections, required more care during pruning and rebuilding to maintain performance.

    \item \textbf{Pruning enables adaptive deployment.} By combining structured pruning, selective weight freezing, and staged rebuilding, we create a single model capable of supporting multiple capacity configurations. This means that a model can adjust to different resource budgets (e.g., edge devices vs servers) without retraining from scratch, making it highly adaptive and efficient at runtime.
\end{itemize}

In summary, the evaluation confirms that pruning is not just a tool for compression, but a foundation for adaptivity. By carefully selecting pruning strategies and combining them with reconstruction and selective fine-tuning, models can be made adaptive. This is key to deploying neural networks efficiently across a range of devices, from high-performance servers to resource-constrained edge systems.

\section{Challenges Encountered}

While building and evaluating the adaptive pruning framework, several challenges came up, both technical and experimental.

\begin{itemize}

    \item \textbf{ResNet’s Skip Connections Were Tricky:} Pruning ResNet-20 was particularly challenging because of its skip (shortcut) connections. These connections require that the number of channels matches between the main and shortcut paths. This meant that pruning one layer often forced us to adjust other layers too, making pruning and rebuilding more complex than in sequential models like VGG.

    \item \textbf{Rebuilding the Model Was Technically Complex:} Restoring pruned weights during the rebuilding process required tracking pruning metadata like indices and tensor shapes. Getting this right across layers especially after several pruning steps was difficult and took time to debug and validate.

    \item \textbf{Training Stability at High Pruning Levels:} When pruning more than 70\% of a model, training became unstable. In some cases, the model couldn't recover accuracy even with fine-tuning. We had to experiment with different learning rates, schedulers, and optimizers to stabilize training after aggressive pruning.

\end{itemize}

\section{Limitations}

While this thesis presents a promising approach to adaptive structured pruning and rebuilding, there are a few limitations. First, the experiments were conducted only on three CNN architectures: VGG-16, AlexNet, and ResNet-20. These models offer a useful range of depth and complexity, but do not include more modern or lightweight architectures like MobileNet, EfficientNet, or transformer-based networks, which limits how broadly the findings can be applied. Additionally, all experiments were performed on the CIFAR-10 dataset, which is relatively small and simple. Larger datasets such as ImageNet might have different pruning behaviors and trends.

We also noticed that the rebuilt models didn’t always perform better than the smaller pruned models, especially when the pruning percentage was low (like 30\%). This suggests that our rebuilding method works best when a larger portion of the model has been pruned. One possible reason is that with more pruning, there's a bigger set of parameters that can be selectively added back and fine-tuned later, which gives the model more room to recover and improve.

Another limitation is that the fine-tuning strategy used across experiments was consistent, but not extensively optimized per pruning level or method. It's possible that more aggressive or adaptive fine-tuning schedules could lead to even better recovery performance.

Lastly, although the framework supports model adaptivity by pruning and rebuilding models offline, it does not yet support dynamic adaptation at runtime. In real-world applications, it would be useful for a model to adapt itself on the fly based on available resources or execution conditions, something that was beyond the scope of this work.

\section{Future Work}
While this thesis presents a practical framework for structured pruning and adaptive reconstruction of CNNs on CIFAR-10, there are several directions for extending this work.

First, applying the proposed methods to larger-scale datasets such as ImageNet would provide a more robust evaluation of performance and generalizability. ImageNet introduces greater complexity in both image resolution and class diversity, making it a valuable benchmark for testing scalability and robustness of pruning strategies.

Second, future work could explore dynamic or data-driven pruning schedules. Instead of selecting fixed pruning percentages (30\%, 50\%, 70\%), adaptive mechanisms could determine the optimal pruning rate at each step based on validation performance or hardware constraints.

Third, although this work focuses on CNNs, it would be valuable to investigate the effectiveness of adaptive pruning and rebuilding in other architectures, such as vision transformers or larger ResNet variants. Such models are increasingly used in practical applications and may benefit significantly from structured compression techniques.

Finally, evaluating the proposed methods in real-world deployment settings such as edge devices or mobile platforms could offer insights into practical resource savings, latency improvements, and energy efficiency.

%% file: sections/conclusion.tex
\chapter{Conclusion}\label{ch:conclusion}

This thesis proposed and evaluated an adaptive pruning-and-rebuilding framework to create elastic CNNs suitable for deployment in dynamic and resource-constrained environments. Our thesis showed that structured pruning, particularly when guided by dependency graphs and advanced importance metrics, significantly reduces model complexity without considerable accuracy degradation. Importantly, our novel rebuilding mechanism proved effective in restoring the network's performance by selectively reintroducing pruned filters and leveraging previously learned knowledge.

Through extensive experimentation on the CIFAR-10 dataset across four diverse CNN architectures, VGG-16, AlexNet, ResNet-20, and ResNet-56, we observed that iterative pruning, followed by fine-tuning and selective freezing during rebuilding, delivered better performance compared to one-shot pruning methods. Our ablation studies further highlighted that selective pruning strategies, such as pruning alternate layers, can provide better accuracy retention, highlighting the subtle relationship between pruning strategy and network performance.

Despite the effectiveness demonstrated, this work faces limitations, including evaluations constrained to the CIFAR-10 dataset and specific CNN architectures. Future research should explore more extensive datasets, a broader range of architectures, and dynamic runtime adaptation capabilities. Overall, this research contributes a comprehensive methodology for developing adaptive CNNs, significantly advancing the potential for efficient and flexible deep learning deployment across varying resource environments.